\documentclass[10pt,twocolumn,letterpaper]{article}

\usepackage[pagenumbers]{cvpr} %

\usepackage{graphicx}
\usepackage{amssymb}

\usepackage[utf8]{inputenc} %
\usepackage[T1]{fontenc}    %
\usepackage{url}            %
\usepackage{booktabs}       %
\usepackage{amsfonts}       %
\usepackage{nicefrac}       %
\usepackage{amssymb}
\usepackage{microtype}      %
\usepackage{graphicx}
\usepackage{booktabs}
\usepackage[normalem]{ulem}
\usepackage{multirow}
\usepackage{amsmath}
\usepackage{rotating}
\usepackage{makecell}
\usepackage{stfloats}
\usepackage{wrapfig}
\usepackage{enumitem}
\usepackage{scalerel}
\usepackage{color}
\usepackage{gensymb}
\usepackage{bm}
\usepackage[dvipsnames]{xcolor}
\usepackage{scalerel}
\usepackage{capt-of}
\usepackage{lineno}
\usepackage{float}

\definecolor{cvprblue}{rgb}{0.21,0.49,0.74}
\usepackage[pagebackref,breaklinks,colorlinks,citecolor=cvprblue]{hyperref}

\DeclareFontFamily{U}{mathb}{}
\DeclareFontShape{U}{mathb}{m}{n}{
  <-5.5> mathb5
  <5.5-6.5> mathb6
  <6.5-7.5> mathb7
  <7.5-8.5> mathb8
  <8.5-9.5> mathb9
  <9.5-11.5> mathb10
  <11.5-> mathb12
}{}
\DeclareSymbolFont{mathb}{U}{mathb}{m}{n}
\DeclareMathSymbol{\drsh}{3}{mathb}{"EB}

\newcommand\blfootnote[1]{%
  \begingroup
  \renewcommand\thefootnote{}\footnote{#1}%
  \addtocounter{footnote}{-1}%
  \endgroup
}

\definecolor{citecolor}{HTML}{0071bc}
\definecolor{gtred}{HTML}{FF3E30}
\definecolor{predblue}{HTML}{0776FF}
\hypersetup{
  breaklinks   = true,
  colorlinks   = true, %
  urlcolor     = RubineRed, %
  linkcolor    = RubineRed, %
  citecolor    = citecolor %
}

\newcounter{extendedfigure}
\renewcommand{\theextendedfigure}{\arabic{extendedfigure}}

\usepackage{caption}
\newcommand{\extendedfigurecaption}[1]{
    \refstepcounter{extendedfigure}
    \captionsetup{labelformat=empty} %
    \caption{Extended Data Figure \theextendedfigure: #1} %
    \captionsetup{labelformat=default} %
}

\def\paperID{*****} %
\def\confName{CVPR}

\title{MatchAnything: Universal Cross-Modality Image Matching with Large-Scale Pre-Training}

\author{
    Xingyi He$^{1}$
    \quad Hao Yu$^{1,2}$
    \quad Sida Peng$^{1}$
    \quad Dongli Tan$^{1}$
    \quad Zehong Shen$^{1}$
    \quad Hujun Bao$^{1\dagger}$
    \quad Xiaowei Zhou$^{1\dagger}$
    \vspace{1em}
    \\
    $^1$State Key Lab of CAD\&CG, Zhejiang University \quad 
    $^2$Shandong University \quad
}

\begin{document}
\maketitle

\blfootnote{$^\dagger$Corresponding authors: Hujun Bao and Xiaowei Zhou.}

\begin{abstract}
Image matching, which aims to identify corresponding pixel locations between images, is crucial in a wide range of scientific disciplines, aiding in image registration, fusion, and analysis. 
In recent years, deep learning-based image matching algorithms have dramatically outperformed humans in rapidly and accurately finding large amounts of correspondences.
However, when dealing with images captured under different imaging modalities that result in significant appearance changes, the performance of these algorithms often deteriorates due to the scarcity of annotated cross-modal training data.
This limitation hinders applications in various fields that rely on multiple image modalities to obtain complementary information.
To address this challenge, we propose a large-scale pre-training framework that utilizes synthetic cross-modal training signals, incorporating diverse data from various sources, to train models to recognize and match fundamental structures across images.
This capability is transferable to real-world, unseen cross-modality image matching tasks.
Our key finding is that the matching model trained with our framework achieves remarkable generalizability across more than \textbf{eight unseen} cross-modality registration tasks using the \textbf{same network weight}, substantially outperforming existing methods, whether designed for generalization or tailored for specific tasks.
This advancement significantly enhances the applicability of image matching technologies across various scientific disciplines and paves the way for new applications in multi-modality human and artificial intelligence~(AI) analysis and beyond.
Project page: \url{https://zju3dv.github.io/MatchAnything/}.
\end{abstract}

\section{Introduction}\label{sec:intro}

\begin{figure*}[htb]
    \centering
    \includegraphics[width=1.0\linewidth]{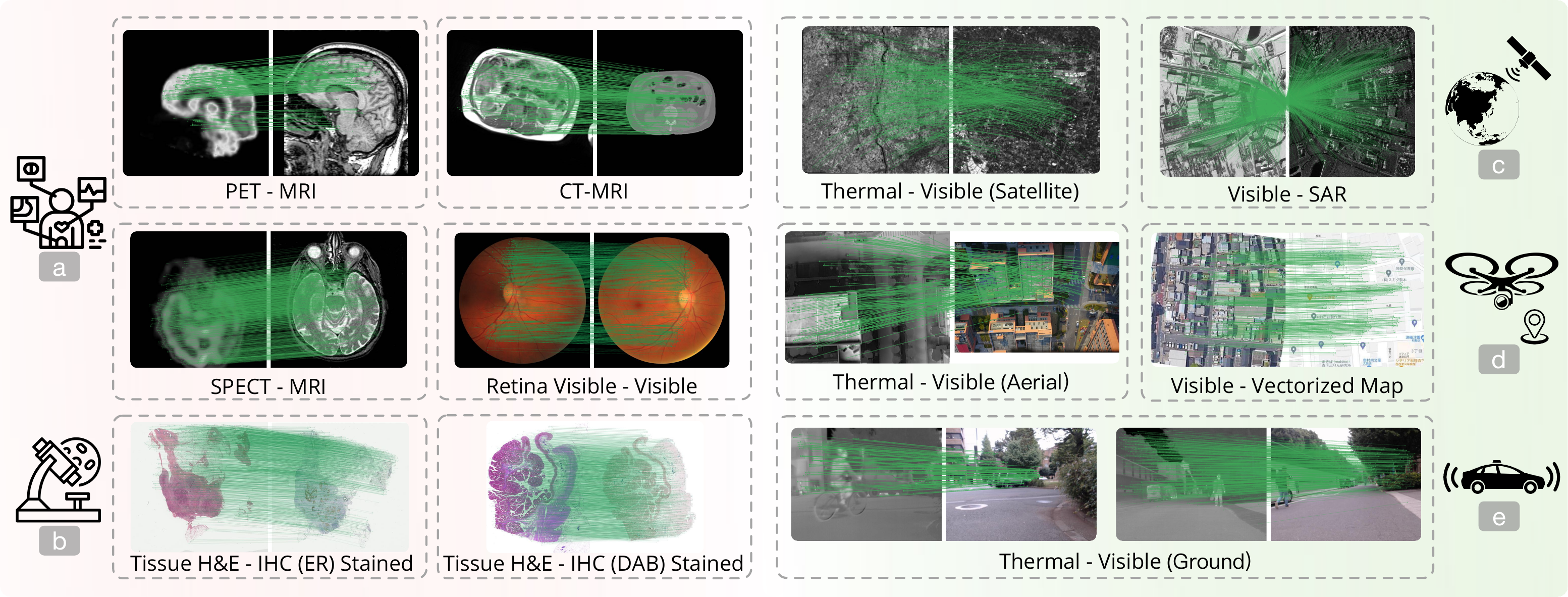}
    \caption{\textbf{Capabilities of the image matching model pre-trained by our framework.} 
    Green lines indicate the identified corresponding pixel localizations between images.
    Using \textbf{the same network weight}, the detector-free matcher~\cite{wang2024eloftr} with Transformer exhibits impressive generalization abilities across extensive unseen real-world single- and cross-modality matching tasks, benefiting diverse applications in disciplines such as (a) medical image analysis, (b) histopathology, (c) remote sensing, autonomous systems including (d) UAV positioning, (e) autonomous driving, and more.
    \textit{The figure is best viewed in color and with zoom-in for clarity.}
    }
    \vspace{-0.35 cm}
    \label{fig:teaser}
\end{figure*}

Cross-modality image matching, which seeks to accurately find corresponding pixel locations between images from different imaging principles, is a fundamental challenge across various disciplines.
The estimated matches serve as the cornerstone for recovering image transformations in image registration, benefiting a wide range of applications in medical image analysis, histopathology, remote sensing, vision-based autonomous systems, and more, as shown in Fig.~\ref{fig:teaser}.
Concretely, in medical image analysis, aligning tomographic images such as MRI with CT, PET, or SPECT from the same patient allows for the combination of different types of information into a single, unified analysis, offering complementary insights and aiding in more accurate diagnosis~\cite{Tempany2015MultimodalIF,Castellano2024AutomatedDO,Song2021AnEM}.
Additionally, in histopathology, matching and registering images of tissue sections stained with different techniques, such as Hematoxylin \& Eosin (H\&E) and various Immunohistochemistry (IHC) stains, which highlight distinct tissue features, can significantly facilitate comprehensive clinical evaluation by human experts or artificial intelligence~\cite{Morrison2021ConventionalHA,Lin2023HighplexII,Pati2024AcceleratingHW}.
In remote sensing, image registration requires matching images captured by different sensors, such as visible light with Synthetic Aperture Radar (SAR) or thermal images, each providing unique imaging advantages.
This registration enables image fusion for multi-modalities image analysis~\cite{Guo2023SkySenseAM,Sharma2021YOLOrsOD}, benefiting applications such as geological exploration, disaster relief, etc.
In vision-based autonomous systems, matching images captured by different sensors, such as visible light and thermal images, enhances robust localization and navigation in low-light environments for applications such as Unmanned Aerial Vehicles (UAVs), autonomous driving, and robotics.

Finding correspondences by human labeling is time-consuming and labor-intensive, making it impractical for processing large datasets.
For example, annotating matches for the registration of 481 histopathology image pairs required approximately 250 hours of work from 9 experts~\cite{Borovec2020ANHIRAN}.
Therefore, many computer vision algorithms have been proposed to address this challenge.
Image matching is originally formulated as a keypoint detection, description, and matching pipeline in a handcrafted or deep learning-based approach. 
Given two input images, a set of salient keypoints is first detected in each image. Local descriptors are then extracted from the neighborhood regions around these keypoints. Finally, corresponding points are identified through nearest neighbor searching in the feature space or by using more sophisticated matching algorithms.
However, cross-modality matching tasks present significant challenges to image matching methods, primarily due to the substantial appearance changes resulting from differences in imaging principles.
The methods~\cite{detone2018superpoint,sarlin20superglue,lindenberger2023lightglue,chen2021learning, shi2022clustergnn} that rely on keypoint detectors often struggle to identify reliable keypoints across images from different modalities in the initial stage, which hinders the subsequent matching process.

Recently, some learning-based detector-free methods~\cite{sun2021loftr,chen2022aspanformer,wang2022matchformer,wang2024eloftr,edstedt2023roma} directly match image pixels with the help of the Transformer~\cite{Vaswani2017AttentionIA} mechanism.
Without the restriction of keypoint detection, they have shown to be more robust than detector-based methods in challenging data such as low-textured scenes and large-perspective changes.
Nevertheless, state-of-the-art detector-free matching methods~\cite{wang2024eloftr,edstedt2023roma}, which are typically trained on abundant single-modality data, exhibit limited generalization to cross-modality tasks.
Additionally, training matching models with strong generalizability for each cross-modality registration task is challenging, as these methods typically require large-scale datasets with dense ground truth correspondences, where the annotated data for cross-modality matching is scarce.
This issue is particularly pronounced in medical research due to patient privacy protections~\cite{2024AQO}.
These limitations hinder the practical applications of cross-modal matching in real-world scenarios.

In this paper, we propose a large-scale cross-modality matching pre-training framework, that can unleash the generalizability of transformer-based detector-free matchers on various unseen real-world cross-modality tasks in different fields.
We identify the main limitation as the lack of annotated cross-modal training data, where there are two critical components to address this problem: (1) \emph{the cross-modal stimulus signals}, which encourage the network to learn appearance-insensitive, fundamental image structural information, thus facilitating generalization to unseen cross-modal tasks.
To achieve this, we propose leveraging pixel-aligned image translation networks~\cite{CycleGAN2017,depthanything} to synthesize images in other modalities for constructing cross-modal training pairs with significant appearance and structural changes.
(2) \emph{The diversity of training data sources}, which serves as the cornerstone for enabling the network to generalize to never-seen-before structures such as satellite views and tissue slices.
Specifically, we employ a mixed training approach that incorporates various types of resources, including multi-view images with scene reconstructions, extensive unlabelled video sequences, and large-scale single-image datasets.
Furthermore, for training on unlabelled video sequences, we devise a coarse-to-fine strategy for constructing pseudo ground truth matches by exploiting the continuity of video frames.
By joint training on these diverse resources, we harness their unique characteristics to improve the robustness and generalization capabilities of image matching networks for unseen cross-modality matching tasks.

The proposed pre-training framework can be effectively applied to several detector-free matching methods without any modifications.
We select ROMA~\cite{edstedt2023roma}, a dense matcher that emphasizes robustness, and ELoFTR~\cite{wang2024eloftr}, a semi-dense matcher that balances both efficiency and effectiveness, as base models for training.
Our findings show that models pre-trained using the proposed framework can exhibit universal generalizability across more than \emph{eight unseen} real-world cross-modal tasks using \emph{the single network weight} without requiring further training, benefiting a wide range of disciplines including medical image analysis, histopathology, remote sensing, autonomous systems, etc.
An overview of our models' capabilities is shown in Fig.~\ref{fig:teaser}.
Extensive experiments conducted on nine datasets reveal that models pre-trained with our framework significantly outperform state-of-the-art matching and image alignment methods in multi-modality registration tasks, whether they are designed for generalization or tailored for specific tasks.
We believe these advances in matching and registration will pave the way for new applications in human and artificial intelligence~(AI)~\cite{Guo2023SkySenseAM,Chen2024TowardsAG,Xu2024AWF} analysis using multi-modality data across various disciplines.

\begin{figure*}[tp]
    \centering
    \includegraphics[width=1.0\linewidth]{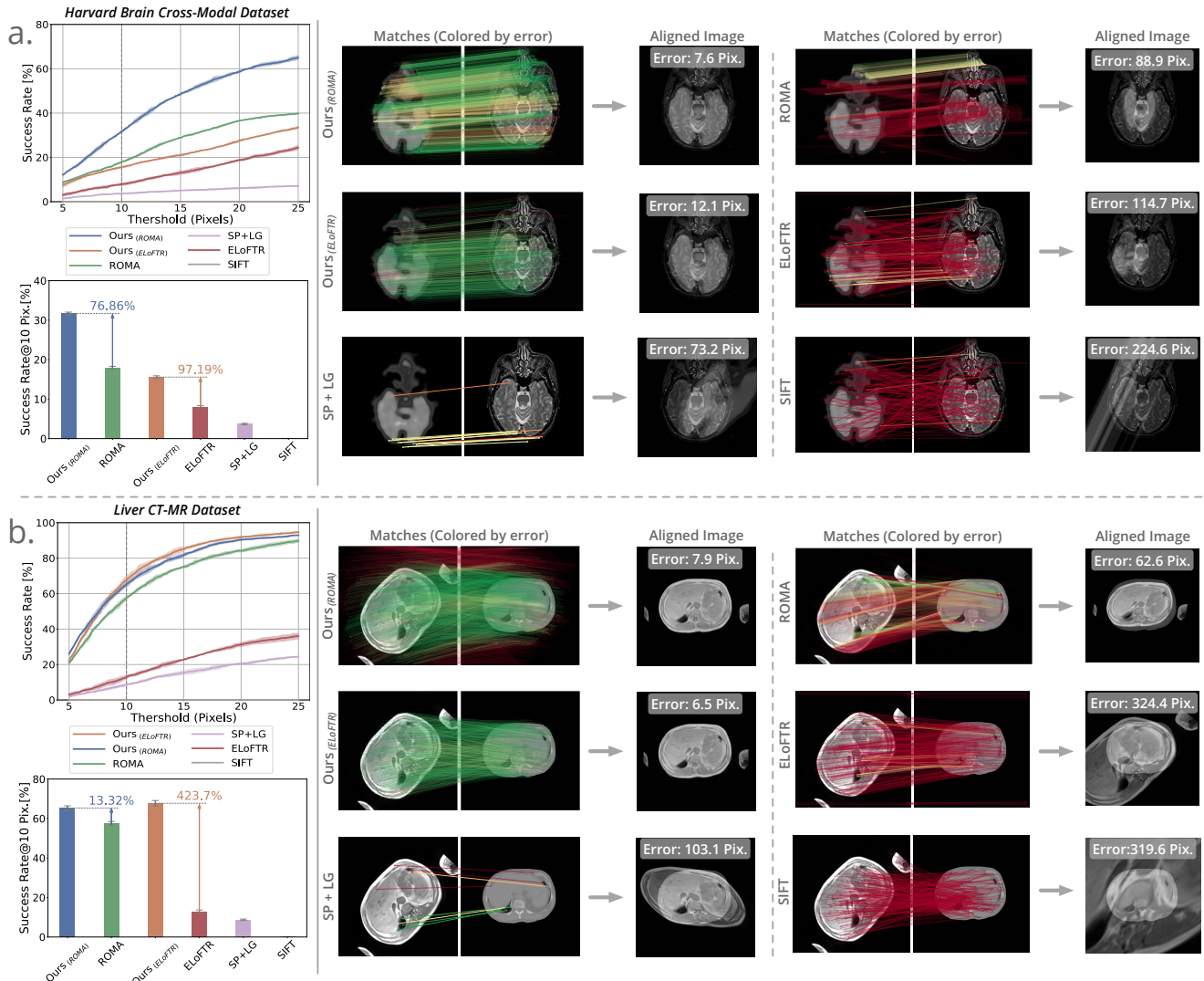}
    \vspace{-0.8cm}
    \caption{\textbf{Comparisons on cross-modality tomography image registration datasets.}
    Our trained models are compared with four representative baselines.
    \textbf{Parts a} and \textbf{b} are the results of the Harvard Brain dataset and Liver CT-MR dataset respectively.
    The curve of the success rate~(SR) metric under different thresholds is shown on the left-up side of each part, where the detailed comparisons with relative improvements on SR@10 pixels metric are shown on the left-down side.
    Qualitative comparisons of predicted matches and aligned images are shown on the right side of each section.
    The matches are colored by the match error, where the green color means that the match error is less than 5 pixels.
    For a full table of quantitative comparisons with baselines, see the Extended Data Tab. ~\ref{tab:exp full}.
    }
    \vspace{-0.5cm}
    \label{fig:ctmrresults}
\end{figure*}

\section{Results}\label{sec:results}
Extensive experiments are conducted to compare the performance of the matching models trained by the proposed large-scale pre-training framework with state-of-the-art image matching and registration methods across nine datasets, encompassing more than eight cross-modality registration tasks.
These tasks span various fields including medical image analysis, histopathology, remote sensing, UAV positioning, and autonomous driving.
We select semi-dense matcher ELoFTR~\cite{wang2024eloftr} and dense matcher ROMA~\cite{edstedt2023roma} for pre-training.
For each model, we use \emph{\textbf{the single weight}} pre-trained by our framework to conduct experiments on \emph{\textbf{all datasets}} without any fine-tuning to illustrate the universal matching capabilities.
For more comprehensive evaluations, we assess the effectiveness of cross-modality matching using metrics specific to downstream tasks, including registration accuracy and pose estimation accuracy.
The details about the experimental settings and metrics are provided in Methods (Sec.~\ref{subsec:expdetails}).

\paragraph{Multi-Modality Tomography Image Registration.}
In medical image analysis and diagnostics, multiple modality images are often used to provide complementary information about the patient's condition.
The commonly used modalities in clinical diagnostics include Computed Tomography (CT), Magnetic Resonance~(MR) imaging, Positron Emission Tomography (PET), and Single Photon Emission Computed Tomography (SPECT).
CT provides rapid, high-resolution images, making it ideal for emergency situations, trauma assessment, and detailed visualization of bone and lung structures.
MR offers superior soft tissue contrast without ionizing radiation, making it the preferred choice for neurological, musculoskeletal, and cardiovascular imaging.
PET excels in detecting metabolic and functional abnormalities, which are crucial for oncology, neurology, and cardiology.
SPECT, while having lower spatial resolution than CT, is valuable for functional imaging in cardiology and neurology due to its cost-effectiveness and lower radiation exposure.
By registering and fusing these images from a single patient, healthcare professionals can combine the strengths of both modalities to make more accurate diagnoses~\cite{Tempany2015MultimodalIF,Castellano2024AutomatedDO,Song2021AnEM}.
Our experiments encompass cross-modality registration tasks involving 2D slices between various modalities.
Specifically, we use the Harvard Brain dataset~\cite{summers2003harvard}, which includes CT-MR, PET-MR, and SPECT-MR brain images from 810 patients, and the Liver CT-MR dataset~\cite{Bauer2020GenerationOA}, which comprises the CT and MR liver images from 111 patients.

The quantitative and qualitative comparisons are shown in Fig.~\ref{fig:ctmrresults}.
Due to the severe appearance and structural differences between images, such as those from SPECT and MR, existing matching matchers often fail to produce accurate matches, leading to erroneous image registration and fusion.
In contrast, the models end-to-end trained by the proposed large-scale pre-training framework achieve significantly better performances than all baselines, despite having never been exposed to medical images or tomography modalities.
Notably, the ROMA~\cite{edstedt2023roma} model trained with our framework shows a 76.9\% relative improvement on the Harvard Brain datasets~(Fig.~\ref{fig:ctmrresults}a), and the ELoFTR~\cite{wang2024eloftr} model trained with our framework achieves a remarkable 423.7\% relative improvement on the Liver CT-MR datasets~(Fig.~\ref{fig:ctmrresults}b).
The considerably improved model generalizability on completely unseen structures and modalities highlights the effectiveness of our training framework, which successfully teaches the model to learn and match fundamental image structures.

\begin{figure*}[tp]
    \centering
    \includegraphics[width=0.95\linewidth]{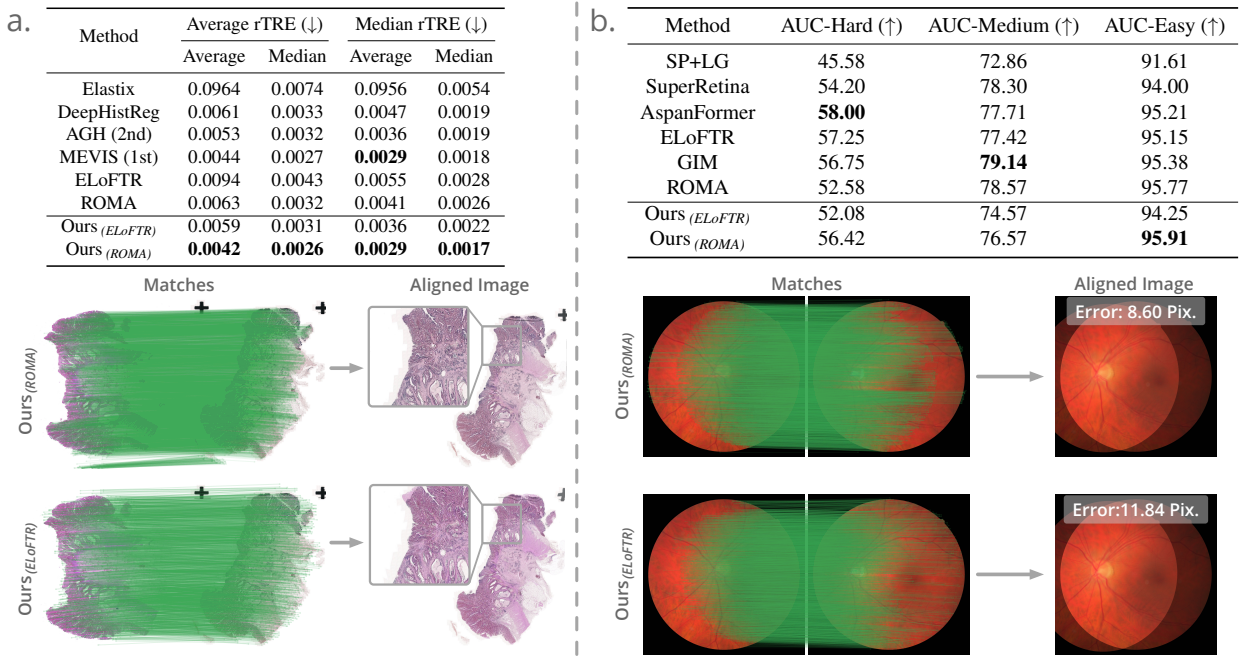}
    \vspace{-0.1cm}
    \caption{\textbf{Results on cross-modality histology registration and retina image registration tasks.}
    \textbf{Parts a, b} show the results of the histology registration task evaluated on the ANHIR dataset and the results of the retina image registration task using the FIRE dataset.
    For each part, the upper table shows the quantitative comparisons with state-of-the-art baselines, whereas the lower figure shows the matching and registration results of our trained models.
    }
    \label{fig:anhirresults}
\end{figure*}

\paragraph{Registration of Histology Images with Different Stains.}
In histopathology, tissue sections are commonly analyzed through histology images employing various staining technologies, such as Hematoxylin and Eosin (H\&E), Immunohistochemistry (IHC), etc.
Each staining method highlights different cellular structures and features.
By registering and fusing histology images from different stains, a more comprehensive understanding of the tissue's pathology can be achieved, facilitating more accurate analysis and diagnoses~\cite{Morrison2021ConventionalHA,Lin2023HighplexII}.
However, the significant appearance variance between different stain methods, and relative displacements and deformations between tissue slices impose considerable challenges to image matching and registration algorithms.
We utilize the ANHIR~\cite{Borovec2020ANHIRAN} challenge dataset to evaluate the performance of cross-modality matching in the registration of histology images stained differently.
This dataset includes tissue sections from various organs including lungs, kidneys, breasts, mammary glands, and more.
It comprises numerous image pairs across different stains, including PAS-CD31, PAS-aSMA, CD1a-CD68, H\&E-Ki67, H\&E-ER, H\&E-PR, etc.

As shown in Fig.~\ref{fig:anhirresults}a, the ROMA model trained by the proposed framework achieves competitive performances on all metrics compared with existing well-engineered optimization-based methods, including the first and follow-up competition solutions.
When compared with the learning-based image alignment method DeepHistReg~\cite{Wodzinski2020DeepHistRegUD}, which directly regresses the warping field and is specially trained for histology registration, our method achieves significantly better accuracy, particularly on the average metrics.
Furthermore, our trained ROMA model shows a $33.2\%$ relative improvement in accuracy over the original ROMA model on the Average-Average rTRE metric, while our trained ELoFTR model achieves a $55.3\%$ relative improvement over its original model.
These results demonstrate that our framework effectively enhances the generalization capabilities of these matching models on previously unseen tissue slices and stains.
\begin{figure*}[tp]
    \centering
    \vspace{-0.7cm}
    \includegraphics[width=1.0\linewidth]{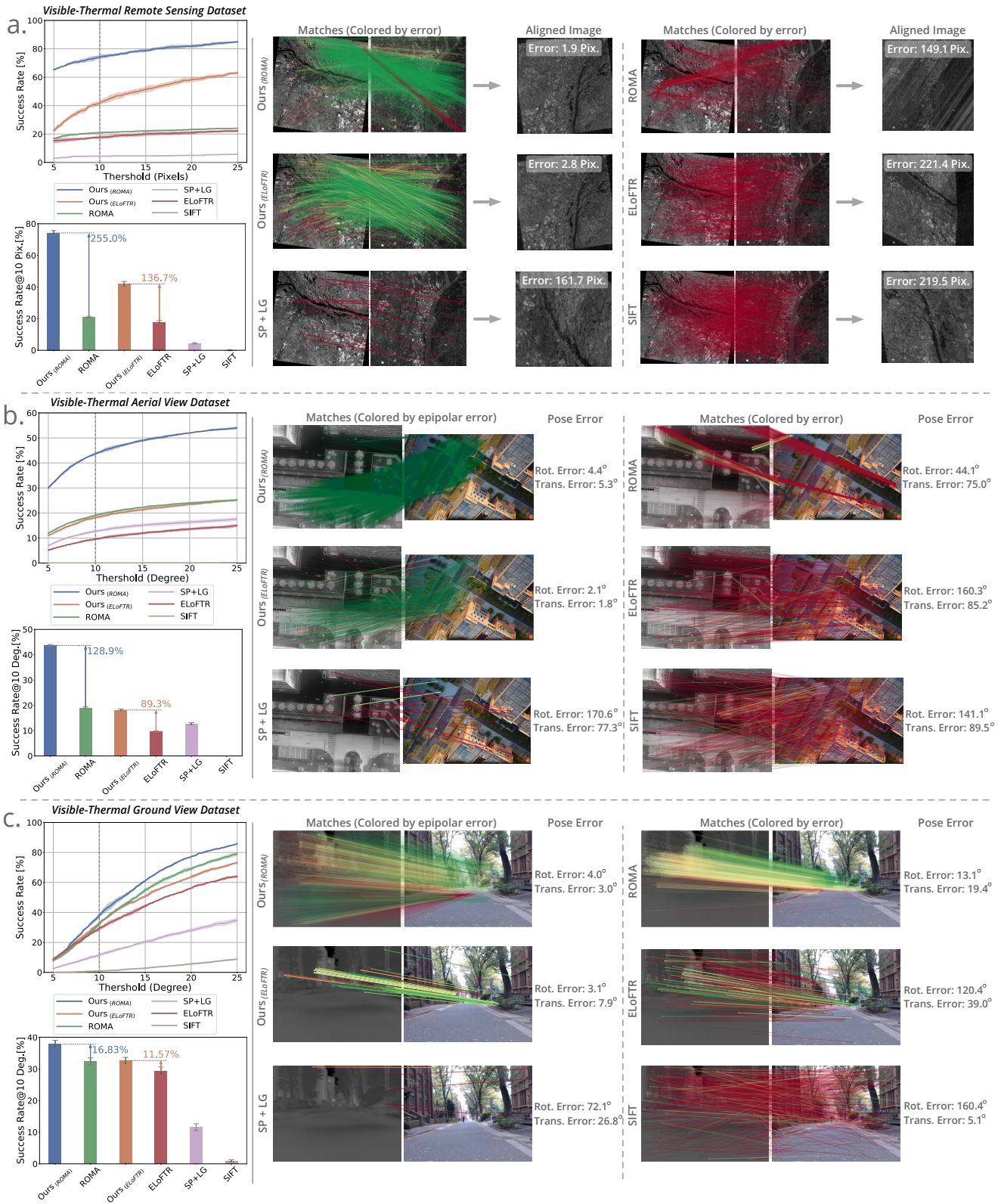}
    \vspace{-0.7cm}
    \caption{\textbf{Results on visible-thermal registration tasks.}
    Our trained models are compared with four representative baselines.
    \textbf{Parts a, b, c} show the results on remote sensing, aerial view, and ground view scenes respectively.
    The left column of each part shows the quantitative comparisons with baselines using success rates~(SR) with a range of thresholds, as well as the detailed comparisons with relative improvements using SR@10 pixels for Part a, and SR@10$\degree$ for \textbf{Parts b, c}.
    The right column shows the qualitative comparisons with baselines in terms of matching quality and registration error.
    The green matches mean the match errors are less than 5 pixels for Part a and epipolar error is less than $3 \times 10^{-3}$ for \textbf{Parts b, c}.
    The aligned images and warping errors are shown in Part a and the pose estimation errors are shown in \textbf{Part b, c}.
    For a full table of quantitative comparisons with baselines, see Extended Data Tab.~\ref{tab:exp full}.
    }
    \label{fig:infraredresults}
\end{figure*}

\paragraph{Retina Image Registration.}
The matching and registration of retina images from different viewpoints play a crucial role in ophthalmology.
The integrated perspectives can aid in the diagnosis and treatment of various eye diseases, such as glaucoma, macular degeneration, and diabetic retinopathy~\cite{Dai2021ADL,Dai2024ADL}.
We use the FIRE~\cite{HernandezMatas2017FIREFI} dataset that contains 134 pairs of visible retina images from different perspectives for evaluation.
This experiment validates the performance of the cross-modality matching models trained by our framework for single-modality tasks.

Results are shown in Fig.~\ref{fig:anhirresults}b.
Existing learning-based matchers perform well on the single-modality visible light matching task. Our findings show that, despite being endowed with strong cross-modality matching capabilities, our models can also achieve comparable performance on visible light matching task. Specifically, our trained ROMA model achieves the best accuracy on the AUC-Easy metric, surpassing SuperRetina~\cite{Liu2022SemiSupervisedKD}, which is specifically trained for retina image matching. These results suggest that cross-modality trained models can serve as general-purpose matchers for single-modality tasks.

\paragraph{Thermal and Visible Light Image Registration.}
Thermal sensors play a crucial role in a wide range of applications, including remote sensing, autonomous systems such as UAV positioning, autonomous driving, robotics, and more.
It captures images that reveal temperature distribution, which is valuable in remote sensing applications~\cite{Guo2023SkySenseAM,Sharma2021YOLOrsOD} such as monitoring mountain forest fires and analyzing urban heat distribution.
Additionally, the ability of thermal sensor to operate effectively in low-light conditions, as well as through fog and smoke, makes them a valuable tool for enhancing the robustness of localization in autonomous systems.
In these applications, matching between thermal and visible light images is essential for tasks such as multi-modal image fusion in remote sensing, multi-sensor calibration in autonomous driving, and visual localization, where thermal images from autonomous devices are matched with maps constructed from visible light data for UAV and robotic navigation.

We evaluate the accuracy of in-plane transformation and relative 6-DoF pose estimation across various thermal and visible light datasets, representing different types of scenes.
Results are shown in Fig.~\ref{fig:infraredresults} and Extended Data Tab.~\ref{tab:exp full}.
SuperFusion~\cite{Tang2022SuperFusionAV}, which is specifically trained for visible-thermal matching task, generalizes poorly on these out-of-distribution datasets.
The models trained with our framework consistently outperform all baseline methods by a large margin across all datasets.
On the Visible-Thermal Remote Sensing dataset~\cite{Li2023MultimodalIM}, which consists of satellite view images, existing matchers perform poorly due to the significantly different characteristics of thermal and visible light images, as well as the considerable difference in perspective from typical training data. 
Our trained ROMA and ELoFTR models achieve success rates (SR) of $74.2\%$ and $41.9\%$ at the 10-pixel metric, respectively, representing relative improvements of $255.0\%$ and $136.7\%$ compared to their original versions.
The Visible-Thermal Aerial View dataset~\cite{Liu2022AMT}, captured from UAV devices, presents challenges due to significant 3D viewpoint changes between images. Our training framework results in relative improvements of $128.9\%$ for the ROMA model and $89.3\%$ for the ELoFTR model.
On the Visible-Infrared Ground View dataset~\cite{Karasawa2017MultispectralOD}, which depicts outdoor street scenes, existing methods perform relatively well due to the similar perspectives in their training data. Nonetheless, our framework still achieves relative improvements of $16.8\%$ for ROMA and $11.6\%$ for ELoFTR on the SR@10\degree metric.
These results consistently demonstrate that our training framework significantly improves models' performance across all these datasets.
Furthermore, they highlight that even though synthetic thermal pairs were used during training, the trained models generalize effectively to real-world thermal-visible data.

\begin{figure*}[tp]
    \centering
    \includegraphics[width=1.0\linewidth]{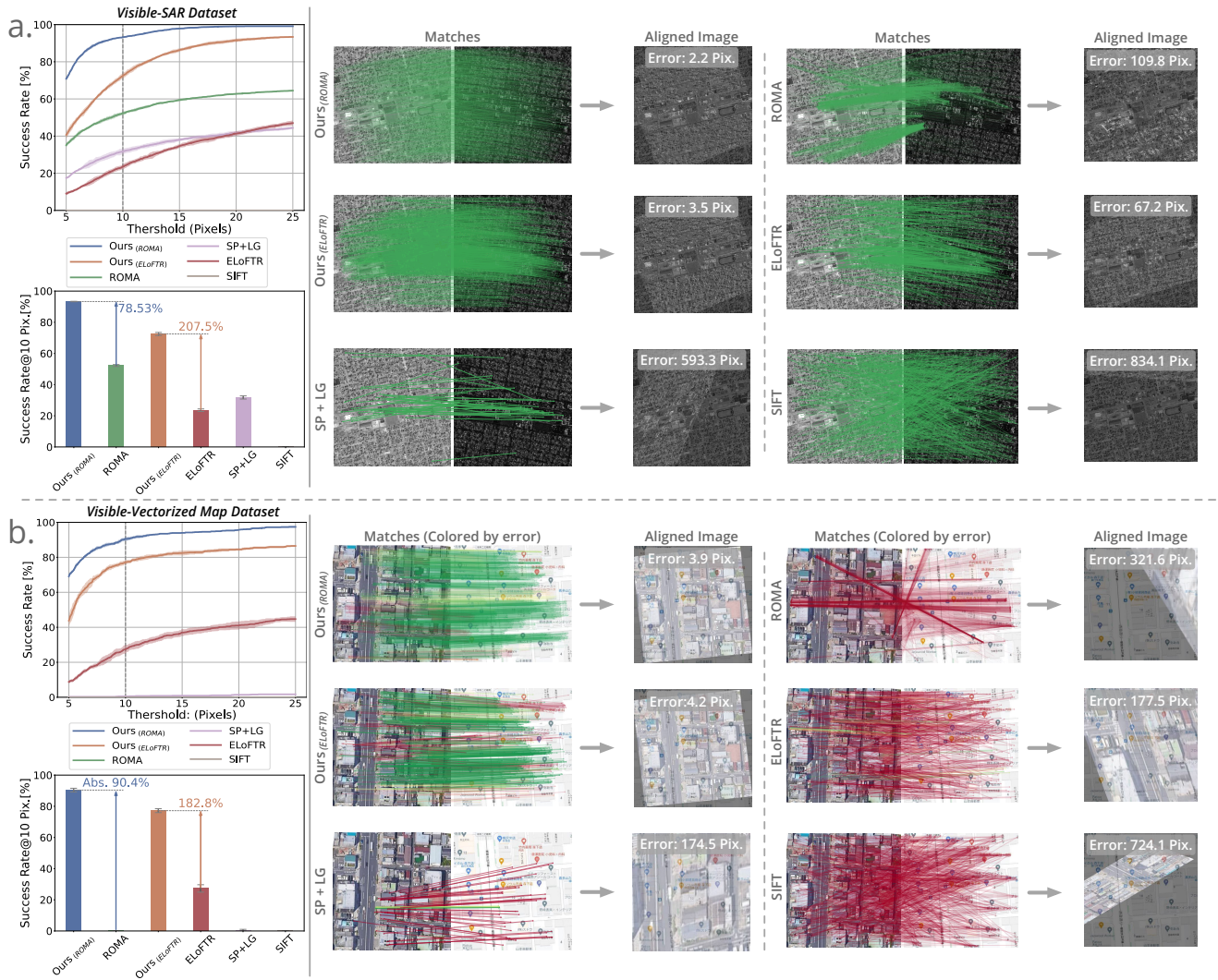}
    \vspace{-0.5cm}
    \caption{\textbf{Results on visible-SAR and visible-vectorized map registration tasks} are shown in \textbf{Parts a, b} respectively.
    Our trained models are compared with four representative baselines.
    The left column shows the quantitative comparisons with baselines using success rate~(SR) metrics at a range of thresholds, as well as the detailed comparisons using SR@10 pixels with the relative improvements of our methods over baselines.
    The right column compares the matching quality and images aligned by the transformations recovered from matches.
    For \textbf{Part b}, the matches are colored by the match errors, where the green means the error is within 5 pixels.
    For a full table of quantitative comparisons with baselines, see Extended Data Tab.~\ref{tab:exp full}.
    }
    \label{fig:sarmapeventresults}
\end{figure*}

\paragraph{SAR Image and Visible Light Image Registration.}
Synthetic Aperture Radar (SAR) images are captured in the microwave portion of the electromagnetic spectrum, allowing them to penetrate clouds, and smoke, providing consistent imaging capabilities regardless of weather conditions or time of day.
This all-weather, day-and-night operational capability ensures reliable data acquisition, which is important for applications in remote sensing such as environmental monitoring, and disaster response.
The matching and registration between SAR and visible light images are essential for achieving information fusion, which significantly enhances comprehensive remote sensing analysis~\cite{Guo2023SkySenseAM}.

We use the Visible-SAR~\cite{xiang2023global} dataset, which contains SAR and visible light image pairs captured from satellite views with perspective changes, and the results are shown in Fig.~\ref{fig:sarmapeventresults}a. The models trained with our framework achieve remarkable performance improvements compared to the baseline methods, despite the SAR modality and satellite viewpoints being completely unseen in the training data. Specifically, our trained ROMA and ELoFTR models achieve success rates of $93.3\%$ and $72.5\%$, respectively, on the 10-pixel metric, representing relative improvements of $78.5\%$ and $207.5\%$ compared to their original versions.

\paragraph{Vectorized Map and Visible Light Image Registration.}
The vectorized map, a commonly used and easily accessible data source, represents the highly abstract layout of urban buildings and is frequently utilized for everyday localization and navigation. Matching and registering vectorized maps with visible light images significantly benefit UAV localization and navigation~\cite{Goforth2019GPSDeniedUL,Zhong2023AHV}, especially in urban scenarios where GPS signals may be obstructed by buildings. This process enhances the self-localization capabilities of devices.
We use \cite{Li2023MultimodalIM} dataset for evaluating the capability of matchers on this task.

Results are shown in Fig.~\ref{fig:sarmapeventresults}b and Extended Data Tab.~\ref{tab:exp full}.
Due to the significant differences in appearance, the existing state-of-the-art matcher ELoFTR performs poorly, and ROMA even completely fails. 
MCNet~\cite{zhu2024mcnet}, which is specifically trained on visible-vectorized map pairs, performs poorly on the out-of-distribution test set.
In contrast, the matching models trained with our framework achieve substantial improvements. As shown in the qualitative results, the trained models produce much better match quality and better image alignment with much smaller errors. Specifically, the SR@10 pixels metric achieves $90.4\%$ for ROMA and $77.2\%$ for ELoFTR.
While SuperFusion~\cite{Tang2022SuperFusionAV} performs well on the SR@20 pixels metric, likely due to its semantic-level supervision mechanism, it performs poorly on high-accuracy metrics.
Conversely, the models trained with our framework achieve significantly better accuracy on the strict SR@5 pixels threshold.
These significant improvements on \emph{completely unseen} image matching tasks between vectorized maps and visible light images highlight the efficacy of our training framework in enhancing both the generalizability and accuracy of matching models.

\paragraph{Running Time.}
We use the same hyperparameters as the original implementations of the ROMA and ELoFTR models, with only the network weights differing. Therefore, the running time of the matching models trained with our framework is the same as the original models. For matching two images with a resolution of $640 \times 480$, the ELoFTR model consumes $40$ms, while the ROMA model takes $303$ms. The running times were evaluated on a single NVIDIA RTX 3090~GPU.

\section{Discussion}\label{sec:discussion}

Cross-modality matching serves as the foundation for multi-modality image registration, which is an important task across various scientific disciplines such as medical imaging, histopathology, remote sensing, autonomous systems, etc.
However, the limited generalizability of existing matching models hinders their practical applications.
In this paper, we introduce a large-scale pre-training framework that enables state-of-the-art detector-free matchers to achieve universal cross-modality matching capabilities on a wide range of unseen tasks.
Our approach begins with a mixed training strategy that incorporates various training data resources, including multi-view image datasets with ground truth reconstructions, extensive unlabelled video sequences, and large-scale single-image datasets.
To effectively utilize unlabelled video data for training, we innovate a coarse-to-fine strategy for constructing pseudo ground truth matches.
This joint training approach leverages the complementary strengths of different datasets to provide rich and diverse training data.
Additionally, we introduce cross-modal stimulus training signals using image generation techniques to encourage the matching model to learn to match appearance-insensitive, fundamental image structures.
Extensive experiments across nine datasets demonstrate that models pre-trained with our framework exhibit remarkable generalizability on more than eight unseen real-world cross-modality registration tasks without requiring additional task-specific training, significantly outperforming state-of-the-art matching and image alignment methods.
We believe these results represent a milestone in computer vision and machine intelligence, paving the way for new applications in human and artificial intelligence (AI)~\cite{Guo2023SkySenseAM,Chen2024TowardsAG,Xu2024AWF} analysis using multi-modality data across a broad range of disciplines.

The limitation of our training framework is that the trained models currently perform poorly on cross-modality matching between aerial view and ground view images, due to the extreme differences in both perspective and appearance.
We believe the issue stems from the lack of relevant training data, as our framework cannot effectively mimic the drastic perspective changes between these views.
This limitation can be addressed in future works by fine-tuning our pre-trained models on specific cross-modality tasks using small-scale labeled data with memory-efficient techniques such as LoRA~\cite{Hu2021LoRALA} and ControlNet~\cite{zhang2023adding}, which can enhance models' task-specific performance while maintaining strong generalization ability.

\section{Methods}\label{sec:method}
\begin{figure*}[t]
    \centering
    \includegraphics[width=0.95\linewidth]{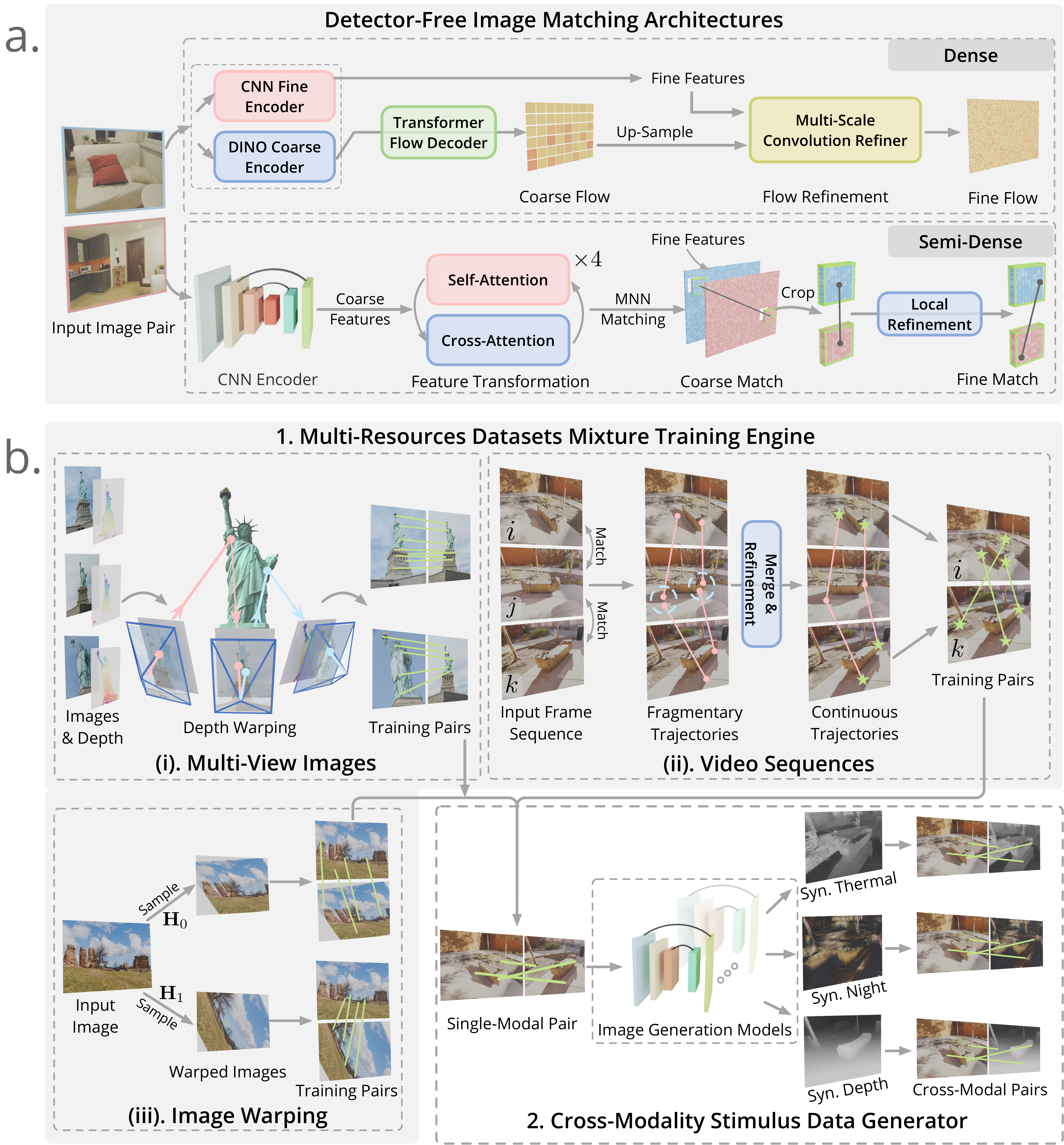}
    \caption{\textbf{Method Overview.} \textbf{a.} We first introduce two types of transformer-based detector-free matching architectures, including dense and semi-dense, serving as base models for our pre-training framework. \textbf{b.} The proposed large-scale universal cross-modality pre-training framework consists of \textbf{(1)} a multi-resource dataset mixture engine designed to generate image pairs with ground truth matches by integrating the strengths of various data types. This engine is composed of (i) multi-view images with known geometry datasets that obtain ground truth matches by warping pixels using depth maps to other images; (ii) video sequences by leveraging the continuity inherent in video frames to construct point trajectories in a coarse-to-fine manner, and then build training pairs with pseudo ground truth matches between distant frames;
    (iii) image warping that sample transformations to construct synthetic image pairs with perspective changes for large-scale single image datasets.
    \textbf{(2)} Subsequently, cross-modality training pairs are generated to train matching models in learning fundamental image structure and geometric information, which is achieved by using image generation models to obtain pixel-aligned images in other modalities, and then substituted for the original image in the training pairs.}

    \vspace{-0.35 cm}
    \label{fig:main}
\end{figure*}

Our objective is to train highly generalizable detector-free image matching models capable of finding accurate correspondences from a pair of images from different unseen modalities during training.
To this end, we propose a large-scale cross-modality pre-training framework that integrates multi-data-resource and multi-modal stimulus signals, as illustrated in Fig.~\ref{fig:main}.
In the following sections, we first give preliminaries about the detector-free matching architectures used by our framework in Sec.~\ref{subsec:train}, and then elaborate on the two key components of our framework: the multi-resource datasets mixture training engine (Sec.~\ref{subsec:joint_train}), and the cross-modal stimulus data generator (Sec.~\ref{subsec:data_gen}).
Related works are reviewed in Sec.~\ref{subsec:related_works}, and experimental details are presented in Sec.~\ref{subsec:expdetails}.
Lastly, ablation studies (Sec.~\ref{subsec:ablation}) are conducted to explore the key design choices of our framework. 

\subsection{Preliminaries about Detector-Free Matchers}\label{subsec:train}

Detector-free matchers, which are end-to-end trained with ground truth matches, have shown remarkable performance with the help of transformer architecture in tackling extreme perspective changes in common single-modality image matching tasks.
We leverage the proposed large-scale, multi-resource, cross-modality training framework to unleash the universal capabilities of detector-free matchers in cross-modality matching tasks.

Our framework can be applied to several detector-free matcher without necessitating modifications to the methods.
In this work, we select two state-of-the-art detector-free matchers as base models for training to demonstrate the efficacy of the proposed framework: ROMA~\cite{edstedt2023roma}, a dense method that focuses on robustness, and ELoFTR~\cite{wang2024eloftr}, a semi-dense method that balances both efficiency and effectiveness.

\subsubsection{Dense Matcher: ROMA}\label{subsec:roma}
ROMA~\cite{edstedt2023roma} predicts a dense warping field $W^{l\rightarrow r}$ given an image pair (marked as left $l$ and right $r$), which is subsequently sampled by considering both match confidence and spatial distributions to extract reliable correspondences.
It firstly extracts coarse image feature maps from input image pairs using a pre-trained DINOv2~\cite{Oquab2023DINOv2LR} backbone and fine features using a CNN backbone.
Then, it predicts the coarse warping field using coarse features using a transformer-based decoder.
Subsequently, the fine warping field in high resolution is achieved by iteratively refining the previous level of the warping field by convolution refinement network using fine-level features.

Due to the expansive parameter spaces in both the encoder and decoder, ROMA exhibits strong advantages in matching scenes with challenging appearance and perspective changes.
This feature makes it particularly well-suited for large-scale pre-training and generalization to unseen cross-modality tasks.

\subsubsection{Semi-Dense Matcher: ELoFTR}\label{subsec:eloftr}
ELoFTR~\cite{wang2024eloftr} achieves detector-free matching through a coarse-to-fine strategy with a transformer mechanism.
Initially, it extracts coarse features in downsampled resolution ($\nicefrac{1}{8}$) and fine features in original image resolution using a CNN network and then applies global self- and cross-attention mechanisms to transform the coarse feature maps for better discriminativeness.
Following this, coarse matches are established through dense mutual-nearest-neighbor~(MNN) matching using these transformed coarse features.
Subsequently, these coarse matches are refined for sub-pixel accuracy: for a coarse match, its corresponding local feature patch in the left and right fine feature maps are cropped and then matched locally by feature correlation and MNN to get an intermediate fine match in pixel-level accuracy.
Finally, it is further refined by keeping the match in the left image fixed while performing feature correlation and expectation on the local $3\times 3$ window around the match in the right image to get the final sub-pixel-level fine match.
As it performs dense pixel-wised matching on downsampled coarse feature maps and then refines it for high accuracy, this scheme is commonly called \emph{semi-dense matching}.

Although it underperforms than ROMA, ELoFTR's advantage is its significantly better efficiency, which is due to its relatively lighter architecture.
This makes ELoFTR especially suitable for applications where speed and computational resource efficiency are critical.

\subsection{Multi-Resources Data Mixture Training}\label{subsec:joint_train}
Providing diverse training data is the cornerstone for training cross-modality matchers with high generalizable capabilities.
The training of the above matching models requires dense ground truth correspondences across images as supervision signals, which are often obtained by warping each pixel in one image to another by known depth values and camera parameters.
However, this process requires access to ground truth reconstruction for each scene, which is expensive to acquire and barricades the scaling up of training data.

To solve this problem, we propose an approach that utilizes joint training across multiple resource datasets, including 
\textit{multi-view images} with known geometry, which are hard to acquire and have limited diversity but offer realistic viewpoint changes along with ground truths; 
\textit{video sequences}, which provide moderately diverse and realistic viewpoint changes but lack ground truth correspondences; 
\textit{image warping} of large-scale single-image datasets, which offers the most diverse but least realistic viewpoint changes.
The joint training of these datasets allows us to leverage the strengths of each dataset while mitigating their individual weaknesses.

\subsubsection{Multi-View Images with Geometry}
With known scene reconstructions, the ground truth correspondences for each image pair can be obtained through depth warping, where a 2D point in the left image is firstly lifted to 3D space in the world coordinates by its depth value and camera parameters, and then projected to the right image by the right camera parameters.

Due to the presence of noise in scene reconstructions, such as inaccurate depth values or scanned meshes with holes, we carefully check warped depth errors and cycle consistency errors to filter out inaccurate ground truth matches.
The warped depth error $e_d$ and cycle projection error $e_c$ are defined as follows:

\vspace{-0.3cm}
\begin{equation*}
    \begin{cases}
        e_d = \frac{\| \*D_{r}(\*x_{proj}) - d_{proj} \|}{\*D_{r}(\*x_{proj})} \enspace, \\
        e_c = \| \*x_l -  \boldsymbol{\pi}_l \cdot \boldsymbol{\xi}_{l \rightarrow r}^{-1} \cdot \*D_{r}(\*x_{proj}) \cdot \boldsymbol{\pi}^{-1}_r (\*x_{proj}) \| \enspace,
    \end{cases}
\end{equation*}
\begin{equation*}
    \text{where} \enspace \*x_{proj}= \boldsymbol{\pi}_r \cdot \boldsymbol{\xi}_{l \rightarrow r} \cdot \*D_l(\*x_l) \cdot \boldsymbol{\pi}^{-1}_l(\*x_l) \enspace.
\end{equation*}
$\*x_l$ is a sampled 2D point in left view, $\*D_{(\cdot)}$ is the depth map of reference or query view, $\boldsymbol{\pi}$ is the projection determined by intrinsic parameters, and $\boldsymbol{\xi}_{l \rightarrow r} = \boldsymbol{\xi}_{r} \cdot \boldsymbol{\xi}_l^{-1}$ is the relative pose between the left and right view. $d_{proj}$ is the $z$ value of 3D points in query view corresponding to $\*x_{proj}$.
A pair of corresponding points is retained in the ground truth matches if the projection depth error $e_d < 0.05$ and cycle projection error $e_c < 3$ pixels.

For this type of data, we use MegaDepth~\cite{li2018megadepth}, ScanNet++~\cite{Yeshwanth2023ScanNetAH}, and BlendedMVS~\cite{Yao2019BlendedMVSAL} datasets for mixture training, encompassing 1079 scenes that cover both indoor and outdoor environments.
However, due to the challenge of collecting high-quality reconstruction data, scaling up this type of dataset is difficult.
Therefore, we incorporate additional diverse datasets including video sequences and single-image data for large-scale training.

\subsubsection{Video Sequences}
Video sequences can be easily collected and provide realistic perspective changes, which are ideal for training image matchers.
However, current large-scale video sequence datasets often lack dense reconstructions for training. 
A promising approach~\cite{xuelun2024gim} involves using state-of-the-art detector-free matchers to match simpler adjacent frames and build trajectories, then leveraging the continuity of video sequences to obtain matches for more distant frames, which are subsequently used as training data.
Nevertheless, since matches produced by detector-free matchers are dependent on image pairs, applying them to consecutive frames leads to inconsistencies, resulting in fragmentary trajectories.
This fragmentation hinders the construction of long-range point tracks, which are essential for obtaining matches for distant pairs with significant perspective changes.
To overcome this problem, we propose a coarse-to-fine strategy to produce long-range tracks and accurate pseudo ground truth matches.
An overview of our approach is shown in Extended Data Fig.~\ref{fig:methoddetail}.

Initially, we sequentially match the $i^{th}$ image with the next $10$ images using ROMA~\cite{edstedt2023roma} model.
To construct point tracks for obtaining matches between distant frames, such as the $i^{th}$ image and the $(i + 40)^{th}$ image that has challenging perspective changes, fragmentary matches are merged to construct coarse trajectories.
Concretely, for each image, we gather all its correspondences with other images as well as the associated confidence values produced by the matcher. 
Then, a non-maximum suppression process is performed across the entire image using a sliding window of size $7\times 7$.
This process merges fragmented matches to locations with the highest confidence within their local region, thus constructing continuous trajectories.

However, the accuracy of correspondences is significantly compromised due to the movement of points during the previous merge process.
To rectify this issue, we further refine the merged point trajectories to achieve sub-pixel accuracy using a transformer-based multi-view refinement approach~\cite{he2024dfsfm}.
For a coarse trajectory, feature patches surrounding the points in each trajectory are first extracted and processed through a multi-view transformer to obtain discriminative features. Subsequently, a dense grid of query points in one view is sampled, and their features are correlated with the feature patches from other views to generate match distribution maps. The sum of variance across all distribution maps is calculated for each query point as the uncertainty criterion. The optimal query point and its peak feature correlation responses in other views are then identified as the refined trajectory locations.

With refined long-range point trajectories, we are able to construct image pairs with pseudo ground truth matches by selecting image pairs that are more than $10$ frames apart and exhibit at least $300$ co-visible correspondences.
In practice, we adopt the DL3DV~\cite{ling2023dl3dv} dataset as the large-scale video dataset for training, which comprises 10K high-quality video sequences, encompassing a broad range of scene categories.
For each pair, 10K matches with high confidence are sampled from the dense warp flow estimated by ROMA to ensure matching quality.
This is followed by geometric verification using RANSAC~\cite{Fischler1981RandomSC} to further remove outliers.

\subsubsection{Image Warping}
Single-image datasets are recognized for their diversity due to the ease of collection from the Internet. They are incorporated into our training framework to further enhance data diversity. 
To create a training pair from a single input image, we apply image warping by sampling homography transformations, which involve adjustments in rotation, translation, scaling, and shearing.
The sampled transformation warps the input image to generate the target image, and the two are then constructed as an image pair.
The corresponding ground truth matches are established by applying the sampled transformation to warp the dense pixel locations in the input image.

However, since this method only creates image pairs with planar transformations, it falls short of accurately mimicking real-world perspective changes.
As a result, models trained exclusively on such warped single-image data often underperform in real-world applications.
To address this issue, we propose to combine single-image warping data with multi-view data with geometry and video sequences in a mixture training manner.
This approach harnesses the diversity of the single-image data while benefiting from the realistic perspective changes provided by multi-view and video data.
In practice, we use large-scale single-image datasets, including GoogleLandmark~\cite{weyand2020GLDv2} and SA-1B~\cite{Kirillov2023SegmentA} for training.

\subsection{Cross-Modality Stimulus Data Generation}\label{subsec:data_gen}
Based on the proposed multi-resources dataset mixture training framework, we now inject the cross-modality stimulus data into the training data, encouraging the network to learn fundamental structural information of images. 
This approach enables the network to generalize effectively on various never-seen-before cross-modality matching tasks.
We propose to use image generation techniques to create synthetic multi-modality image pairs for training.

Concretely, we transform one image from each training pair to other modalities using pixel-aligned image generalization models.
The generated images are then substituted for the original images in the training pairs to form new cross-modal pairs.
The key aspect of this approach is the pixel-aligned property of the image generation models, as it enables the obtained images to maintain the same structural information as the input image while exhibiting significantly different appearances.
This alignment allows for using correspondences from original pairs in training.
In practice, we use the image style translation network and monocular depth estimation network as pixel-aligned image generation models.

\paragraph{Image Style Translation} is a generative model that translates the style of one image to another while preserving the content.
We use this technique to generate images in different modalities with significant appearance changes for training.
Notably, we do not expect the network to ``memorize'' a specifically trained cross-modality matching task, as synthetic models cannot fully replicate real-world imaging principles, such as thermal images that depict temperature distributions in a scene.
Instead, we use synthetic image pairs with significant appearance changes to train the network to learn to match fundamental image structures across different modalities. This learned capability can then be transferred to real-world cross-modality tasks.

Specifically, we train image translation models separately to translate visible light images into thermal and nighttime images, which are two common imaging conditions in real-world scenarios.
Due to the limited resources of pixel-aligned ground truth image pairs for training translation model, we use CycleGAN~\cite{CycleGAN2017} as the image style translator, which can be trained in an unsupervised manner using cycle consistency loss and demonstrates strong generalization capabilities.
In practice, the visible light to infrared translation model is trained on the Tardal~\cite{liu2022target} dataset, and the daytime-nighttime translation model is trained on the Aachen~\cite{Zhang2020ReferencePG} dataset.

\paragraph{Monocular Depth Estimation.} 
We find that previous image style translation models primarily mimic the appearance changes between different modalities, with limited impact on image structure. However, real-world modality differences extend beyond mere appearance variations. For example, regions with sufficient visual texture in visible light images may lack texture when captured by a thermal sensor due to minimal temperature differences.

To address this, we introduce image structural changes by incorporating visible light and depth map pairs into the training process.
Monocular depth estimators predict the depth value of each pixel in the input image and have shown remarkable generalization ability based on large-scale pre-training.
Moreover, depth maps introduce significantly larger structural changes compared to images generated through style translation, enhancing the matching models' abilities to handle cross-modal variations.
We experimented with various depth estimation networks and selected DepthAnything~\cite{depthanything} considering its optimal balance of efficiency and performance.
The estimated depth maps are rescaled to grayscale images, which then replace the original images to form cross-modal pairs.
Notably, for datasets with ground-truth geometry, their depth maps are directly used to generate cross-modality pairs, bypassing the need for depth estimation networks.

\subsection{Training Details}\label{subsec:impl}
Our multi-resource cross-modality training framework generates $\sim$800M image pairs, comprising visible-visible, visible-synthetic thermal, visible-synthetic nighttime, and visible-depth map pairs.
For the matching models, we separately train ROMA and ELoFTR using their official implementations, maintaining identical hyperparameters and loss functions for fair comparisons with their original models.
Training is carried out on 16 NVIDIA A100-80G GPUs with a batch size of 64. The training process takes approximately 4.3 days for ELoFTR and 6 days for ROMA. 
The AdamW optimizer~\cite{Loshchilov2017DecoupledWD} is used with an initial learning rate of $8\times 10^{-3}$. 
For each method, a single pre-trained model weight is used to conduct all experiments presented in this paper, highlighting the strong generalizability unleashed by our training framework.

\subsection{Related Works}\label{subsec:related_works}

\paragraph{Cross-Modality Image Registration.}
Image registration seeks to estimate 2D or 3D transformations between a pair of 2D images.
These transformations include planar transformations such as affine or homography, relative camera poses with six degrees of freedom (6DoF) in 3D space, and non-rigid transformations like B-splines.
The goal of estimating these transformations is to enable the fusion of images or to facilitate camera calibration and localization tasks, which are essential in applications across various fields.
Many traditional methods~\cite{ArgandaCarreras2006ConsistentAE, Avants2008SymmetricDI, Glocker2011DeformableMI, Rueckert1999NonrigidRU, Klein2010elastixAT, Lotz2019RobustFA, Wodzinski2019AutomaticNH} typically adhere to a pipeline where initial image correspondences are established through human labeling or by using a 2D image matching algorithm such as SIFT~\cite{LoweDavid2004DistinctiveIF}.
Following this, the transformation is solved based on these correspondences, allowing for image alignment via the estimated transformation.
Optionally, for enhancing precision, non-rigid alignment is subsequentially employed for image pairs with deformation.
This involves estimating non-rigid B-spline transformations through non-linear optimization, utilizing point warping distance loss and a mattes mutual information~\cite{Mattes2001NonrigidMI} loss, which is intensity-invariant and thus more robust to the varying appearances across different modalities.
A recent matching method SRIF~\cite{Li2023MultimodalIM} extends SIFT to handle multiple cross-modality tasks by proposing an image intensity transformation to address significant appearance changes between modalities.
However, it still relies on handcrafted designs for intensity transformation and matching.
Learned priors can not be used for benefiting matching, limiting its performance.

As for deep-learning-based methods, DeepHistReg~\cite{Wodzinski2020DeepHistRegUD} proposes a neural network to directly regress the deformation field to register histology images with different stains, which is trained in a self-supervised manner.
However, its application is restricted to a single task, and it demonstrates limited accuracy due to a lack of diverse training data.
\cite{Balakrishnan2018VoxelMorphAL,meng2024correlationawarecoarsetofinemlpsdeformable} use 3D convolution or attention mechanisms to regress the deformation field to register brain images of the same patients at different times.
However, these methods are inherently task-specific and struggle with cross-modality data, such as registration between CT and MRI, SPECT and MRI, or when applied to other organs like kidneys, primarily due to the limited availability of training data.
Moreover, their dependency on 3D volumes as input restricts their application to 2D images, limiting their versatility in broader clinical settings.
SuperFusion~\cite{Tang2022SuperFusionAV} regresses dense optical flow for the fusion of visible light and thermal images, trained with semantic supervision.
Nevertheless, its generalizability is limited due to the lack of diversity in the training dataset, and it is not well-suited for matching images with large perspective changes, as it relies on the small displacement assumption of optical flow.
XoFTR~\cite{Tuzcuoglu2024XoFTRCF} builds on LoFTR~\cite{sun2021loftr} with a modified image matching model trained on synthetic thermal data. However, its limited diversity in modality and scene variations constrains its generalizability to other cross-modality matching tasks.
Some methods~\cite{zhu2024mcnet, Zhao2021DeepLH} directly regress the homography transformation for an input image pair, which can be used for cross-modality registration.
However, the application of these methods is limited to planar images and cannot be extended to 6DoF pose or non-rigid transformation estimations.

In this work, we address all these limitations by developing universal cross-modality image matchers that are capable of providing accurate correspondences essential for the transformation estimation pipeline across various tasks.

\paragraph{Image Matching.}
Classical image matching methods~\cite{LoweDavid2004DistinctiveIF, rosten2006machine, bay2008speeded} rely on handcrafted techniques for detecting keypoints, describing them, and then performing matching by nearest-neighbor searching.
In contrast, recent advancements employ deep neural networks for both detection~\cite{rosten2006machine,savinov2017quad,barroso2019key,Liu2022SemiSupervisedKD} and description~\cite{tian2017l2,mishchuk2017working,tian2019sosnet,ebel2019beyond}, significantly enhancing the robustness of detection and discriminativeness of local descriptors.
Additionally, some modern methods~\cite{dusmanu2019d2,detone2018superpoint,revaud2019r2d2,luo2020aslfeat,tian2020d2d} managed to learn the detector and descriptor together.
SuperGlue~\cite{sarlin20superglue} introduces the transformer~\cite{Vaswani2017AttentionIA} mechanism into matching.
Subsequent work LightGlue~\cite{lindenberger2023lightglue} further improves the efficiency and accuracy of the transformer-based matcher SuperGlue using the strategy of adapting to the matching difficulty for early stop matching.

Detector-free methods match images directly, bypassing the need for detecting specific keypoints and instead producing semi-dense or dense matches. This approach enhances robustness and demonstrates greater model capabilities.
The early methods~\cite{rocco2018neighbourhood, rocco2020efficient} achieve this by 4D correlation volumes.
LoFTR~\cite{Shen2022SemiDenseFM} first employs the transformer~\cite{Vaswani2017AttentionIA} in detector-free matching to model long-range dependencies and produce semi-dense matches in a coarse-to-fine manner.
Many follow-up works~\cite{wang2022matchformer, chen2022aspanformer, tang2022quadtree} further improve the matching accuracy, by performing attention on multi-scale features~\cite{wang2022matchformer, chen2022aspanformer}, using the guidance of flow~\cite{chen2022aspanformer}, or restricts the attention span during hierarchical attention to relevant areas~\cite{tang2022quadtree}.
To improve efficiency while preserving robustness, ELoFTR~\cite{wang2024eloftr} proposes an aggregated attention mechanism that adaptively selects tokens and sequences redundant computations.
It achieves significantly higher efficiency than LoFTR while exhibiting better matching accuracy.
Recently, Dense matching methods~\cite{truong2021learning,edstedt2023dkm,edstedt2023roma}, which are designed to estimate all possible correspondences between two images, have shown strong robustness on large perspective changes.
As a side effect, they are generally much slower compared with sparse and semi-dense methods due to the heavier architectures.

However, cross-modality matching tasks remain challenging for these learning-based matchers due to severe appearance changes between images.
With well-developed matching architectures, the bottleneck lies in the lack of large-scale cross-modality training data.
The proposed large-scale pre-training framework circumvents this problem and can benefit multiple learning-based matching methods.

\paragraph{Training of Image Matchers.}
Previous keypoint detection and description method~\cite{detone2018superpoint} employs a multi-stage training strategy, which includes human labeling and self-supervised techniques, due to the ambiguity inherent in defining keypoints.
Matchers~\cite{sarlin20superglue,sun2021loftr,chen2022aspanformer, wang2024eloftr,edstedt2023dkm,edstedt2023roma} typically utilize training datasets composed of multi-view images corresponding scene reconstructions to generate ground truth matches.
Some methods~\cite{sarlin20superglue,lindenberger2023lightglue} also incorporate a pretraining phase that involves single-image warping before training on real-world datasets.
To address the scarcity of data with known reconstructions, CAPS~\cite{Wang2020LearningFD} adopts weak supervision using epipolar geometry constraints.
However, this strategy leads to suboptimal performance compared to a fully supervised approach~\cite{Zhou2020Patch2PixEP} when generalizing to different tasks.
The recent method GIM~\cite{xuelun2024gim} proposes generating ground truth correspondences for unlabelled video sequences through a matching-and-propagation process that leverages the inherent continuity of video. However, it faces challenges with inconsistent correspondences from detector-free matches, necessitating a merging strategy. This method is constrained to using small merge ranges to prevent significant decreases in match accuracy. 
As a result, this limitation restricts the propagation length for constructing image pairs with challenging perspective changes.

Differently, we propose a new cross-modality multi-resources mixture training framework that effectively harnesses the advantages of multi-view images, single-image warping, and video sequences. Additionally, we devise a coarse-to-fine strategy aimed at obtaining long-range and accurate ground truth correspondences for unlabelled video sequences.

\subsection{Experimental Details}\label{subsec:expdetails}
In the following sections, we provide details about the evaluation datasets used in our experiments and baseline settings.
Full results with broader baseline methods as well as using different evaluation metrics are provided in the Extended Data Tab.~\ref{tab:exp full}, ~\ref{tab:expauc}.
We conducted experiments five times with different random seeds, and the average results are reported. Error bars representing the standard deviation are also shown in the figures.
\subsubsection{Evaluation Datasets}
\paragraph{Liver CT-MR Dataset~\cite{Bauer2020GenerationOA}} contains the aligned CT and MR volumes from the same patients.
To construct the evaluation dataset, we uniformly slice five images from each volume and randomly warp images to simulate the misalignments between CT and MR images, which is a common scenario in clinical practice.
There are 555 pairs used for evaluation in total.

Since the image pairs are created by image warping, the ground truth transformation is naturally known and be used for evaluation.
We evaluate the performances of methods by solving affine transformation for each image pair. 
Then, a set of control points $\{\mathbf{x}_s\}$ in the source image are warped by the predicted $\hat{\mathbf{T}}$ and the ground truth $\mathbf{T}_{\text{gt}}$ transformations respectively, whereas the warping error is calculated by the mean Euclidean distance between the two sets of warped points:

\vspace{-0.3cm}
\begin{equation*}
    \text{Error} = \frac{1}{m} \sum_{i=1}^{m}  \| \hat{\mathbf{T}}({\mathbf{x}}^i) - \mathbf{T}_{\text{gt}}({\mathbf{x}}^i) \|_2 \enspace ,
\end{equation*}
where $m$ is the total number of control points.
The registration is considered successful if the warping error is less than $n$ pixels threshold, where the success rate~(SR) over all pairs is reported.
In practice, we use four corners of the source image as the control points.

\paragraph{Harvard Brain Dataset~\cite{summers2003harvard}} contains the accurately aligned brain image pairs between CT-MR, PET-MR, and SPECT-MR, which contains 810 pairs in total.
We create evaluation pairs by randomly warping images, following the same way as the Liver CT-MR dataset.
We use the success rate of mean point warping error under $n$ pixels as the evaluation metric, the same as with the previous Liver CT-MR dataset.

\paragraph{Histology Image Dataset.} We use ANHIR challenge dataset~\cite{Borovec2020ANHIRAN} for evaluation, which covers the tissue section image pairs collected from various organs and conditions including lesions, lungs, mammary glands, Colon Adenocarcinoma (COAD), kidneys, and breasts. It comprises $251$ test pairs across different stains, including H\&E-Ki67, H\&E-ER, H\&E-PR, etc.
A set of ground truth correspondences is provided for each pair, which are annotated and carefully checked by experts.
To evaluate the matching performances on these pairs with non-rigid deformations, we first estimate the rigid affine transformation between each pair, which serves as the rough alignment.
Subsequentially, the fine-grained B-spline non-rigid transformation is optimized by SGD under the constraint of estimated matches.

We follow the evaluation metrics in the ANHIR challenge.
For a pair $(\mathbf{I}_i, \mathbf{I}_j)$ from all test pairs, the evaluation landmark $\mathbf{x}_l^i$ in the source image $\mathbf{I}_i$ is first warped to $\mathbf{I}_j$ by the solved transformations.
Then, the relative Target Registration Error (rTRE) is computed by the Euclidean distance between the warped points $\hat{\mathbf{x}}_l^j$ and the annotated ground truth correspondences $\mathbf{x}_l^j$ normalized by image dialog $d_j$:

\vspace{-0.3cm}
\begin{equation*}
    \text{rTRE}_l^{ij} = \frac{\| \hat{\mathbf{x}}_l^j - \mathbf{x}_l^j \|_2}{d_j} \enspace.
\end{equation*}
Subsequentially, the average rTRE~(ArTRE) and median rTRE~(MrTRE) are computed on all evaluation landmarks in the pair.
The overall performance of the algorithm is evaluated by the average ArTRE, average MrTRE, mean ArTRE and mean MrTRE metrics overall test pairs.

\paragraph{Retina Image Dataset.}
We use the FIRE dataset~\cite{HernandezMatas2017FIREFI} to evaluate the performances of our large-scale cross-modality pre-trained models on retina visible-visible image pairs.
The dataset contains 134 image pairs with manually annotated ground truth correspondences for each pair.
The original paper divided these pairs into easy, moderate, and hard subsets for separate evaluation. 

We follow the evaluation protocol in the SuperRetina~\cite{Liu2022SemiSupervisedKD}.
For each pair, we estimate the homography transformation using the produced correspondences, warp the landmarks from the source image to the target image, and calculate the mean Euclidean distance between warped landmarks and their ground truth matches.
Then, the Area Under the Curve (AUC) of mean warping error at a threshold of $25$ pixels is used as the evaluation metric.

\paragraph{Thermal-Visible Satellite Dataset} compresses 200 image pairs from satellite images.
For each pair, the ground truth transformation is provided in the form of a $3 \times 3$ matrix.
Same with the previous Liver CT-MR dataset, we compute the mean Euclidean distance between the warped control points in the source image by the ground truth transformation and the solved transformation by produced matches.
The success rate~(SR) metrics of warping error under different thresholds are reported.

\paragraph{Thermal-Visible Aerial View Dataset and Ground View Dataset.}
We use \cite{xiang2023global} as an aerial view evaluation dataset, containing 2145 image pairs and depicting real-world 3D perspective changes.
For each pair, the ground truth camera pose is attached.
As for the ground view dataset, we use \cite{Karasawa2017MultispectralOD}.
This dataset provides video image sequences and their corresponding pixel-aligned thermal image sequences.
Since the ground truth camera pose is not provided, we first leverage the current state-of-the-art SfM method~\cite{he2024dfsfm} to recover camera poses using the visible light image sequences.
Due the the pixel alignment between the visible and thermal images, the estimated camera poses are also applicable to the thermal images.
Then, we create thermal and visible image pairs for evaluation, where $492$ pairs are sampled.

The relative pose $(\hat{\mathbf{R}}, \hat{\mathbf{t}})$ between each pair is recovered by solving the essential matrix using produced matches, where the RANSAC~\cite{Fischler1981RandomSC} is utilized to remove outliers.
Then, we calculate the relative pose error using the ground truth camera pose:

\vspace{-0.4cm}
\begin{equation*}
    \text{Error}=\max(R_{\text{err}}, t_{\text{err}}) \enspace,
\end{equation*}
\begin{equation*}
\text{where} \enspace R_{\text{err}} = \arccos \left( \frac{\text{trace}(\hat{\mathbf{R}}^T \mathbf{R}_{\text{gt}}) - 1}{2} \right) \enspace,
\end{equation*}
\begin{equation*}
t_{\text{err}} = \arccos \left( \frac{\hat{\mathbf{t}} \cdot \mathbf{t}_{\text{gt}}}{\|\hat{\mathbf{t}}\| \|\mathbf{t}_{\text{gt}}\|} \right) \enspace.
\end{equation*}
The success rate of the relative pose estimation under various error thresholds is reported.

\paragraph{Visible-SAR Dataset.}
We use the \cite{xiang2023global} dataset that contains 1209 visible-SAR image pairs, where a set of ground truth matches are provided for each pair.
The affine transformation is estimated by predicted matches, which are used to warp the evaluation landmarks from the source image to the target image.
We use the success rate of the mean warping error under different error thresholds as the evaluation metric.

\paragraph{Visible-Vectorized Map Dataset.}
\cite{Li2023MultimodalIM} dataset provides 200 visible-vectorized map image pairs, which are perfectly aligned.
Like the previous Liver CT-MR dataset, we randomly warp the images to create the perspective changes and obtain ground truth transformations.
The success rates of warping errors under different error thresholds are used as metrics.

\subsubsection{Baselines}
\paragraph{Image Matching Baselines.}
We compare our large-scale trained models with handcrafted matching methods SIFT~\cite{LoweDavid2004DistinctiveIF}, SRIF~\cite{Li2023MultimodalIM}, learning-based homography regressor MCNet~\cite{zhu2024mcnet} and a set of state-of-the-art learning-based matching methods, including ROMA~\cite{edstedt2023roma}, DKM~\cite{edstedt2023dkm}, GIM~\cite{xuelun2024gim}, SuperFusion~\cite{Tang2022SuperFusionAV}, ELoFTR~\cite{wang2024eloftr}, MatchFormer~\cite{wang2022matchformer}, AspanFormer~\cite{chen2022aspanformer}, and SuperPoint~\cite{detone2018superpoint}+LightGlue~\cite{sarlin20superglue}~(SP+LG).
For learning-based image matching methods, we use their outdoor models for evaluation. The ROMA, DKM, ELoFTR, MatchFormer, and AspanFormer models are trained on the MegaDepth dataset in a fully supervised manner.
The LightGlue is firstly pre-trained on the Oxford-Pairs dataset~\cite{Radenovic-CVPR18} by single-image homography warping with strong photometric augmentations, including blur, hue, saturation, illumination, etc, and then fine-tuned on the MegaDepth dataset by full supervision.
The GIM model is trained on the proposed video dataset using the video propagation strategy.
Its trained DKM model with optimal performance is used for comparisons.
SuperFusion is trained specifically for visible light image and thermal image matching and fusion using MSRS~\cite{Tang2022PIAFusionAP} dataset.
For MCNet, we utilize the weight trained on visible-vectorized map pairs from the GoogleMap dataset.
Their results are obtained by running open-source code using their released models and hyperparameters.
The SuperRetina~\cite{Liu2022SemiSupervisedKD} baseline is compared on the retina image registration dataset.
It was specially trained on retina datasets in a semi-supervised manner using partially human-labeled retina image pairs and a progressive keypoint expansion strategy.
Its results are from their original paper since the same evaluation dataset and metrics are used in our experiments.

\paragraph{Image Alignment Methods.}
We compare the proposed method with optimization-based and learning-based image registration methods on the non-rigid histology image registration task using the ANHIR challenge dataset. 
For all these methods, we use the results reported by the original papers.
The Elastix~\cite{Klein2010elastixAT} is open-source software that aligns image pairs by optimizing a B-spline deformation using mattes mutual information similarity criteria with the adaptive stochastic gradient descent optimizer.
The first-place solution MEVIS~\cite{Lotz2019RobustFA} and the second-place solution AGH~\cite{Wodzinski2019AutomaticNH} in competition device well-engineered pipelines.
MEVIS first performs initial alignment by trying multiple different image rotations and then estimating affine transformation using the Gaussian-Newton method.
Then non-rigid transformation is found using curvature regularization and L-BFGS optimization.
The second-place solution AGH applies several different approaches and automatically selects the best solution.
It first determines the rigid transformation using RANSAC from matches produced by multiple matching methods including SIFT~\cite{LoweDavid2004DistinctiveIF}, SURF~\cite{Bay2006SURFSU}, and ORB~\cite{Rublee2011ORBAE}.
Then, a non-rigid transformation was found using local affine registration, various versions of the demons algorithms, or a feature-point-based thin-plate spline interpolation.

The DeepHistReg~\cite{Wodzinski2020DeepHistRegUD} baseline directly regresses the rigid affine transformations, as well as the non-rigid deformation fields for the histology image registration task.
It was trained on the ANHIR~\cite{Borovec2020ANHIRAN} training set in an unsupervised manner by minimizing the negative normalized cross-correlation cost function.
Its results are from their original paper since the same evaluation dataset and metrics are used in our experiments.

\subsection{Ablation Studies}\label{subsec:ablation}
We conducted several experiments to validate the design choices of our large-scale pre-training framework using the ROMA matching model on multiple cross-modality evaluation datasets.

\paragraph{Cross-Modality Activation Signals.}
We ablate the incorporation of cross-modality stimulus signals in the training phase.
Firstly, we drop all cross-modality data, where only visible image pairs are used for training.
Extended Data Tab.~\ref{tab:exp ablation}~(1) shows that performances decrease significantly on all the cross-modality evaluation datasets, which indicates the generalizing capability on the unseen modality is greatly reduced.
We also tried to replace the cross-modality data with the commonly used photometric image augmentation methods to bring appearance changes, including illumination, blur, saturation, and hue.
As shown in Extended Data Tab.~\ref{tab:exp ablation}~(2), the photometric augmentation brings little improvements compared with using only visible image pairs.
However, there is a significant performance gap between using the proposed cross-modality stimulus signals.
These experiments validate the efficacy of using cross-modality stimulus data to train image matching models, significantly enhancing their generalizability on unseen cross-modality tasks.

Subsequently, we further analyze the influence of each part of the cross-modality data.
Results in Extended Data Tab.~\ref{tab:exp ablation}~(3) show that the dropping of the synthetic thermal data leads to a 9.7$\%$ drop of SR@10$\degree$ metric on the real-world thermal-visible registration dataset.
The performances on other cross-modality tasks also consistently decreased.
This indicates the use of synthetic thermal training data can not only improve the generalizability of real-world thermal data but also improve the performance on other cross-modality tasks.
Without either depth maps or synthetic night-time images, the performances of trained models decrease, as shown in Extended Data Tab.~\ref{tab:exp ablation}~(4, 5).
Even though these two parts of training modalities are considerably different from the test modalities, they still provide valuable information for image matching models to learn cross-modality matching.

\paragraph{Multi-Resources Dataset Mixture Training.}
We first replace our multi-resources dataset mixture training strategy with a multi-stage manner, which trains on the single image datasets, on the multiview geometry datasets, and video datasets sequentially.
Results reported in Extended Data Tab.~\ref{tab:exp ablation}~(6) illustrate that the multi-stage training with carefully tuned learning rates underperforms the proposed multi-resources dataset mixture training strategy.
We believe the benefits of joint training come from leveraging the best of all worlds from these three data resources.
Then, we drop the single-image datasets and video datasets respectively to see the impact of each data source.
Without either of them, the performances of the trained model drop significantly, as shown in Extended Data Tab.~\ref{tab:exp ablation}~(7, 8).
The results demonstrate the rich diversity of these two data sources is attributed to the generalizability of matching model on unseen image structures and cross-modality tasks.

\paragraph{Coarse-to-Fine Video Dataset Ground Truth Generation.}
Extended Data Tab.~\ref{tab:exp ablation}~(9) reports the results of replacing the proposed coarse-to-fine point trajectory construction strategy by merging the matches with a distance less than 1 pixel to preserve accuracy as in GIM~\cite{xuelun2024gim}.
The proposed strategy brings 7.1$\%$ improvement on SR@5 pixels on the visible-SAR dataset, which demonstrates the effectiveness of the proposed strategy.
These results emphasize that the advantage of our coarse-to-fine strategy is that we can use the larger merge window with a size of $7\times7$ to produce longer trajectories, which attributes to construct image pairs with larger perspective changes.
Then, the following multi-view track refinement phase helps to improve the trajectory accuracy to the sub-pixel level, which is crucial for accurate ground truth training correspondences.

\appendix
\section*{Supplementary Material}
\section{Details about training data generation}\label{supsec:details}
\subsection{Multi-view images with geometry datasets}
\paragraph{MegaDepth~\cite{li2018megadepth}} is an outdoor dataset collected from the Internet with 196 different scenes.
We use the training pairs sampled by LoFTR~\cite{sun2021loftr}.
Notably, the low-quality scenes reported by~\cite{tyszkiewicz2020disk}~(`0000', `0002', `0011', `0020', `0033', `0050', `0103', `0105', `0143', `0176', `0177', `0265', `0366', `0474', `0860', `4541') and scenes that overlap with IMC test set~(`0024', `0021', `0025', `1589', `0019', `0008', `0032', `0063') are removed from training.
\paragraph{ScanNet++~\cite{Yeshwanth2023ScanNetAH} and BlendedMVS~\cite{Yao2019BlendedMVSAL}} contains 380 indoor scenes, 502 indoor and outdoor scenes respectively.
We construct training image pairs by calculating the image overlap ratio $r$ by first warping image pixels from the left image to the right image using the depth maps and camera parameters and then checking depth consistency.
The overlap ratio $r$ is defined as the ratio of the number of pixels that are consistent in warping consistency to the number of pixels in the left image.
For both datasets, we sample the image pairs with an overlap ratio range of $[0.1, 0.7]$

\subsection{Video sequences dataset}
We construct training image pairs with ground-truth matches from unlabeled video sequences.
DL3DV~\cite{ling2023dl3dv} dataset is used, which contains 10K high-quality video sequences from indoor and outdoor scenes.
Firstly, we downsample the video sequences by an interval of 4 frames and then perform image matching between each image and its consecutive $10$ images.
The ROMA model is used in our experiment, where $10$K matches are sampled for each pair by the matching confidences and distributions for the best match quality.
Then, coarse point trajectories are constructed by merging matches using a non-maximum-suppression process with a $7\times 7$ sliding window.
These trajectories are then refined for high accuracy by off-the-shelf multi-view refinement model~\cite{he2024dfsfm}.
The training pairs are constructed by finding image pairs with a distance of more than $20$ frames, containing more than $300$ co-visible matches, and relative mean points motion larger than $30$ pixels.
We use 16 A100-80GB GPUs for processing, which takes about 72 hours to process all the video sequences.

\subsection{Single image datasets}
We use large-scale single image datasets GoogleLandmark~\cite{weyand2020GLDv2} and SA-1B~\cite{Kirillov2023SegmentA} for training.
For each image, random homography transformations are sampled to transform the image to synthesize a new view with perspective changes.
The homography transformation sampling is achieved by combining a uniformly sampled rotation between $[-180\degree, +180\degree]$, translation factor between $[-0.25, +0.25]$, scale factor between $[0.5, 2.0]$, non-isotropic skew factor between $[-0.1, 0.1]$, left-right perspective factor between $[-0.5, 0.5]$, and up-down perspective factor between $[-0.5, 0.5]$.
The ground truth correspondences are obtained by warping dense grid points in the left image to the right image using the sampled homography transformation.

We observe that in this way, the sky parts of outdoor images will also have ground truth matches, where the depth estimation is not reliable in this case.
Directly using these data for training cross-modality image pairs such as between visible light image and estimated depth map will lead to the instability of the training process.
We solve this problem by only supervising the area where the estimated depth map from DepthAnything~\cite{depthanything} is larger than zero, where we find that DepthAnything can naturally filter out the sky parts by depth values.

\section{Experimental details}
For our trained models and baselines ROMA~\cite{edstedt2023roma}, DKM~\cite{edstedt2023dkm}, GIM~\cite{xuelun2024gim}, ELoFTR~\cite{wang2024eloftr}, MatchFormer~\cite{wang2022matchformer}, AspanFormer~\cite{chen2022aspanformer} and SuperPoint~\cite{detone2018superpoint}+LightGlue~\cite{lindenberger2023lightglue}, SIFT~\cite{LoweDavid2004DistinctiveIF}, we feed them resized image pairs with longest edge equal to $840$.
For baseline GIM, we use its best performance model large-scale trained using DKM.
On the histology image registration experiment, the results of baselines Elastix, AGH, and MEVIS are from the ANHIR competition summary paper~\cite{Borovec2020ANHIRAN}.
The OpenCV RANSAC~\cite{Fischler1981RandomSC} is used for filtering out the outliers, where the iteration time is set to $1K$, and the confidence value is set to $0.99999$. 

The Liver CT-MR dataset, Harvard Brain dataset, and Visible-Vectorized map dataset provide pixel-aligned image pairs.
For each pair, we randomly sample transformation and apply it to the right image to obtain evaluation image pairs with perspective changes.
For the Liver CT-MR dataset and Harvard Brain dataset, the transformations are composed of uniformly sampled rotation between $[-50\degree, 50\degree]$, translation factor between $[-0.2, 0.2]$, and scale factor between $[0.75, 1.33]$.
For the visible-vectorized map dataset and Harvard Brain dataset, the transformations are composed of uniformly sampled rotation between $[-10\degree, 10\degree]$, translation factor between $[-0.1, 0.1]$, and scale factor between $[0.8, 1.25]$.

{
   \small
   \bibliographystyle{ieeenat_fullname}
   \bibliography{main}
}

\clearpage
\setcounter{extendedfigure}{0} %
\begin{figure*}[t]
    \centering
    \includegraphics[width=0.9\linewidth]{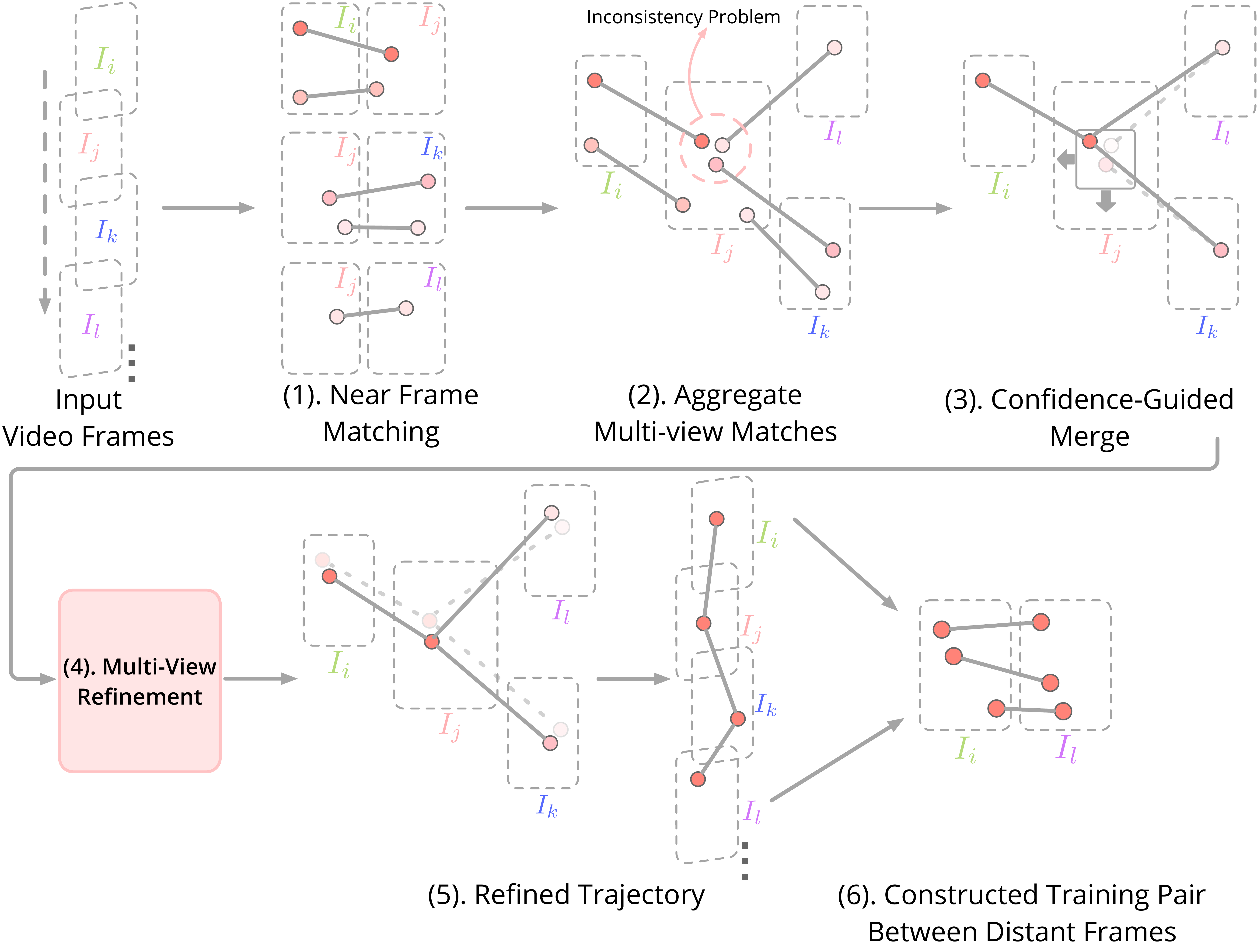}
    \extendedfigurecaption{\textbf{Construct training pairs with ground truth matches for unlabelled video sequences.}
    For training with large-scale unlabelled video sequence datasets, we propose a coarse-to-fine framework to construct ground truth matches between distant image frames with challenging perspective changes by leveraging the continuity of video sequences. 
    (1). Given a set of consecutive frames $\textbf{I}_{i,j,k,l}$, we perform image matching using a state-of-the-art detector-free matcher~\cite{edstedt2023roma} between near frames, which is relatively easy matching task due to the small perspective changes between adjacent frames.
    The confidence scores associated with each match are visually represented by color shades.
    These near-frame matches are used to construct point trajectories, which are crucial for establishing matches between distant frames.
    However, the detector-free matchers are dependent on each image pair, where applying them to multiple images will lead to inconsistencies, resulting in fragmentary trajectories, as highlighted by the red circle in step (2).
    The proposed coarse-to-fine framework addresses this problem by first aggregating all matches and their corresponding confidence scores across frames for each image. For instance, in step (2), we demonstrate the aggregated matches for $\textbf{I}_{j}$.
    (3) Next, we apply a non-maximum suppression process over the frame using a window size of $7\times7$ with matching confidence as a criterion.
    This process can merge the fragmented matches into a single localization with the highest confidence within their local neighborhood, resulting in continuous trajectories.
    Despite obtaining continuous trajectories, the merging process can reduce their accuracy due to point movements.
    To correct this, we perform (4) multi-view refinement of the entire trajectory using a transformer-based approach~\cite{he2024dfsfm}, achieving precise trajectories.
    These refined trajectories allow us to establish accurate matches between distant frames $\textbf{I}_i$ and $\textbf{I}_l$, which serve as ground truths for training.
    }
    \vspace{-0.35 cm}
    \label{fig:methoddetail}
\end{figure*}

\clearpage

\begin{table*}[t]
    \centering
    \resizebox{1.0\textwidth}{!}{
    \setlength\tabcolsep{4pt} %
    \begin{tabular}{lcccccccc}
    \toprule
    \multirow{2}{*}{Method} & \multicolumn{2}{c}{Harvard Brain} & \multicolumn{2}{c}{Visible-Thermal Aerial View} &  \multicolumn{2}{c}{Visible-SAR} &  \multicolumn{2}{c}{Visible-Semantic Map}\\ 
    \cmidrule(lr){2-3}
    \cmidrule(lr){4-5}
    \cmidrule(lr){6-7}
    \cmidrule(lr){8-9}
                  & SR@5pix       & SR@10pix  & SR@5pix   & SR@10pix  & SR@5pix       & SR@10pix & SR@5pix       & SR@10pix \\ 
    \midrule
    Full & \textbf{12.15} & \textbf{31.70} & \textbf{30.01} & 43.65 & \textbf{70.91} & \textbf{93.29} & \textbf{69.10} & \textbf{90.40} \\
    \hline
    (1) w/o all cross-modality signals & 8.86 & 19.96 & 14.29 & 19.98 & 40.01 & 51.29 & 2.20 & 3.00 \\
    (2) replaced by photometric augmentations & 9.29 & 22.01 & 14.79 & 21.87 & 42.11 & 55.62 & 1.60 & 3.50 \\
    (3) w/o synthetic thermal modality & 10.60 & 30.49 & 21.29 & 33.96 & 53.41 & 86.91 & 69.70 & 86.40\\
    (4) w/o synthetic night-time modality & 9.88 & 27.61 & 26.91 & 37.28 & 55.56 & 83.61 & 50.10 & 79.80\\
    (5) w/o depth modality & 10.01 & 25.31 & 22.71 & 35.98 & 64.88 & 88.12 & 40.70 & 65.20 \\
    \hline 
    (6) multi-stages training & 12.10 & 30.17 & 26.89 & 40.01 & 65.83 & 86.19 & 63.10 & 83.40\\
    (7) w/o single-image dataset & 9.37 & 25.87 & 27.80 & \textbf{44.31} & 53.69 & 71.04 & 54.60 & 76.90\\
    (8) w/o video dataset & 8.12 & 27.94 & 24.56 & 33.12 & 57.42 & 74.29 & 61.50 & 80.30\\
    (9) w/o coarse-to-fine design in video dataset & 10.51 & 29.26 & 26.25 & 39.21 & 63.81 & 90.43 & 67.50 & 85.70\\
    \bottomrule
    \end{tabular}
    }

    \caption{\textbf{Ablation Studies.}
    We conduct experiments to validate the design choices of our framework with the ROMA matching model on multiple cross-modality matching tasks. The success rate at different thresholds is used as a metric.
    As shown in (1-5), omitting each component of the cross-modality training signals or replacing them with photometric augmentations consistently degrades performance across all tasks. This demonstrates the effectiveness of using cross-modality training data in enabling the matching model to generalize to unseen cross-modality image pairs.
    (6-8) demonstrate the effectiveness of the multi-resource mixture training strategy. Switching from simultaneous to multi-stage training, or excluding either single-image or video sequence datasets, results in performance degradation.
    Finally, (9) highlights the importance of the proposed coarse-to-fine framework in generating ground truth matches on the video sequence dataset.
    }
    \label{tab:exp ablation}
    \end{table*}

\clearpage

\begin{table*}[t]
    \centering

    \resizebox{1.0\textwidth}{!}{
    \setlength\tabcolsep{20pt} %
    \begin{tabular}{cccccccccc} 
    \toprule
    \multirow{2}{*}{Method}         & \multicolumn{3}{c}{Liver CT-MR Dataset}             & \multicolumn{3}{c}{Harvard Brain Dataset}\\ 
    \cmidrule(lr){2-4}
    \cmidrule(lr){5-7}
                  & SR@5pix       & SR@10pix       & SR@20pix  & SR@5pix       & SR@10pix    & SR@20pix \\ 
    \midrule
    MCNet& 0.00 & 0.00 & 0.04 & 0.00 & 0.22 & 3.43\\
    SIFT& 0.00 & 0.00 & 0.00 & 0.00 & 0.00 & 0.00\\
    SRIF& 1.58 & 9.87 & 29.95 & 1.50 & 6.67 & 11.21\\
    SP+LG& 1.87 & 8.72 & 20.50 & 1.51 & 3.70 & 6.12\\
    AspanFormer& 3.24 & 12.25 & 26.13 & 2.47 & 4.57 & 11.23\\
    MatchFormer& 0.00 & 2.52 & 7.03 & 2.22 & 4.94 & 8.52\\
    ELoFTR & 3.03 & 12.94 & 31.28 & 3.09 & 7.90 & 18.81\\
    SuperFusion& 0.00 & 0.00 & 0.00 & 0.00 & 0.64 & 7.65 \\
    DKM& 6.85 & 23.24 & 44.14 & 6.67 & 11.85 & 23.58\\
    GIM& 16.76 & 46.49 & 75.68 & 4.44 & 8.52 & 13.46\\
    ROMA & 20.94 & 57.62 & 84.22 & 8.67 & 17.93 & 36.59\\
    \hline
    $\text{Ours}_{\textit{ (ELoFTR)}}$ & 22.09 & \textbf{67.75} & \textbf{91.93} & 7.78 & 15.58 & 27.56\\
    $\text{Ours}_{\textit{ (ROMA)}}$ & \textbf{26.05} & 65.30 & 90.38 & \textbf{12.15} & \textbf{31.70} & \textbf{58.72}\\
    \bottomrule
    \end{tabular}
    }

    \vspace{0.5cm}

    \resizebox{1.0\textwidth}{!}{
    \setlength\tabcolsep{10pt} %
    \begin{tabular}{cccccccccc}
    \toprule
    \multirow{2}{*}{Method}         & \multicolumn{3}{c}{Visible-Thermal Remote Sensing  Dataset}             & \multicolumn{3}{c}{Visible-Thermal Aerial View Dataset} & \multicolumn{3}{c}{ Visible-Thermal Ground View Dataset} \\
    \cmidrule(lr){2-4}
    \cmidrule(lr){5-7}
    \cmidrule(lr){8-10}
                  & SR@5pix       & SR@10pix       & SR@20pix  & SR@5$\degree$ & SR@10$\degree$ & SR@20$\degree$  & SR@5$\degree$ & SR@10$\degree$ & SR@20$\degree$\\ 
    \midrule
    MCNet & 1.10 & 1.40 & 5.10 & - & - & - & - & - & - \\
    SIFT & 0.10 & 0.10 & 0.10 & 0.00 & 0.00 & 0.07 & 0.04 & 0.93 & 5.65\\
    SRIF & 1.10 & 5.30 & 11.50 & 0.00 & 0.00 & 0.00 & 0.33 & 1.71 & 5.36\\
    SP+LG& 3.00 & 4.40 & 5.00 & 7.02 & 12.74 & 16.36 & 2.64 & 11.59 & 28.09\\
    AspanFormer& 7.50 & 9.10 & 11.70 & 6.01 & 11.00 & 13.85 & 4.47 & 21.34 & 44.92\\
    MatchFormer & 7.40 & 8.80 & 10.70 & 4.66 & 8.95 & 12.54 & 4.88 & 21.95 & 51.83\\
    ELoFTR & 15.20 & 17.70 & 21.00 & 5.21 & 9.59 & 13.57 & 8.60 & 29.27 & 56.23\\ 
    SuperFusion & 0.90 & 1.50 & 1.50 & 0.00 & 0.13 & 1.08 & 0.00 & 2.15 & 9.80 \\
    DKM& 3.80 & 5.60 & 7.40 & 8.67 & 14.59 & 18.60 & 6.50 & 24.19 & 49.39\\
    GIM& 3.70 & 4.50 & 4.70 & 8.67 & 17.11 & 23.22 & 4.47 & 17.28 & 41.26\\
    ROMA& 17.00 & 20.90 & 23.20 & 11.92 & 19.07 & 24.02 & 7.64 & 32.36 & 69.15\\
    \hline
    $\text{Ours}_{\textit{ (ELoFTR)}}$ & 22.60 & 41.90 & 58.40 & 11.00 & 18.15 & 23.50 & 8.60 & 32.66 & 63.75\\
    $\text{Ours}_{\textit{ (ROMA)}}$ & \textbf{65.30} & \textbf{74.20} & \textbf{81.80} & \textbf{30.01} & \textbf{43.65} & \textbf{52.00} & \textbf{8.79} & \textbf{37.80} & \textbf{77.18}\\
    \bottomrule
    \end{tabular}
    }

    \vspace{0.5cm}

    \resizebox{1.0\textwidth}{!}{
    \setlength\tabcolsep{20pt} %
    \begin{tabular}{cccccccccc}
    \toprule
    \multirow{2}{*}{Method} & \multicolumn{3}{c}{Visible-SAR Dataset} & \multicolumn{3}{c}{Visible-Vectorized Map Dataset} \\
    \cmidrule(lr){2-4}
    \cmidrule(lr){5-7}
                  & SR@5pix       & SR@10pix       & SR@20pix  & SR@5pix       & SR@10pix       & SR@20pix\\ 
    \midrule
    MCNet & 0.00 & 0.00& 0.00 & 0.20 & 1.00 & 12.40\\
    SIFT & 0.00 & 0.00 & 0.00 & 0.00 & 0.00 & 0.00 \\
    SRIF & 0.00 & 0.00 & 0.00 & 0.00 & 0.10 & 0.51\\
    SP+LG& 17.40 & 31.75 & 41.68 & 0.20 & 0.50 & 1.40\\
    AspanFormer & 6.41 & 15.64 & 26.27 & 1.50 & 5.70 & 8.70\\
    MatchFormer& 15.73 & 33.92 & 48.42 & 18.30 & 39.40 & 52.90\\
    ELoFTR & 9.07 & 23.57 & 41.41 & 8.80 & 27.30 & 41.00 \\
    SuperFusion & 0.00 & 0.00 & 0.00 & 19.20 & 67.20 & 95.10 \\
    DKM& 18.01 & 31.72 & 44.99 & 2.30 & 6.50 & 11.30\\
    GIM& 26.19 & 44.82 & 59.84 & 0.00 & 0.00 & 0.00\\
    ROMA& 35.12 & 52.26 & 62.74 & 0.00 & 0.00 & 0.00\\
    \hline
    $\text{Ours}_{\textit{ (ELoFTR)}}$ & 40.69 & 72.47 & 91.56 & 43.80 & 77.20 & 84.50\\
    $\text{Ours}_{\textit{ (ROMA)}}$ & \textbf{70.91} & \textbf{93.29} & \textbf{99.09} & \textbf{69.10} & \textbf{90.40} & \textbf{97.40}\\
    \bottomrule
    \end{tabular}
    }

    \caption{\textbf{Detailed results across multiple datasets.}
    We present statistical comparisons across a full list of baselines, using the success rate~(SR) metric at different thresholds. Compared to homography regression method MCNet~\cite{zhu2024mcnet}, the sparse matching methods SIFT~\cite{LoweDavid2004DistinctiveIF}, SRIF~\cite{Li2023MultimodalIM} and SuperPoint~\cite{detone2018superpoint}+LightGlue~\cite{lindenberger2023lightglue}~(SP+LG), semi-dense matching methods AspanFormer~\cite{chen2022aspanformer}, MatchFormer~\cite{wang2022matchformer}, ELoFTR~\cite{wang2024eloftr}, dense matching methods SuperFusion~\cite{Tang2022SuperFusionAV}, DKM~\cite{edstedt2023dkm}, GKM~\cite{xuelun2024gim}, and ROMA~\cite{edstedt2023roma}, the models trained with our framework outperform them by a large margin on all cross-modality matching and registration tasks. \textbf{Bold} indicates the best performance.
    }
    \label{tab:exp full}
    \end{table*}

\clearpage

\begin{table*}[t]
    \centering

    \resizebox{1.0\textwidth}{!}{
    \setlength\tabcolsep{17pt} %
    \begin{tabular}{cccccccccc} 
    \toprule
    \multirow{2}{*}{Method}         & \multicolumn{3}{c}{Liver CT-MR Dataset}             & \multicolumn{3}{c}{Harvard Brain Dataset}\\ 
    \cmidrule(lr){2-4}
    \cmidrule(lr){5-7}
                  & AUC@5pix       & AUC@10pix       & AUC@20pix  & AUC@5pix       & AUC@10pix    & AUC@20pix \\ 
    \midrule
    MCNet & 0.00 & 0.00 & 0.02 & 0.00 & 0.07 & 0.72\\
    SIFT & 0.00 & 0.00 & 0.00 & 0.00 & 0.00 & 0.00 \\
    SRIF & 0.39 & 2.75 & 11.74 & 0.40 & 2.30 & 5.96\\
    SP+LG& 0.56 & 2.94 & 9.10 & 0.54 & 1.67 & 3.35\\
    AspanFormer& 0.78  & 4.12 &11.95 & 0.69 & 2.02 & 4.79\\
    MatchFormer& 0.00 & 0.60  &2.81 & 1.02 & 2.42 & 4.58\\
    ELoFTR & 0.83 & 4.31 & 13.52 & 0.95 & 3.34 & 8.27 \\
    SuperFusion & 0.00 & 0.00 & 0.00 & 0.00 & 0.13 & 1.80\\
    DKM& 1.90 & 8.62 & 22.05 & 2.19 & 5.78 & 11.95\\
    GIM& 4.56 & 18.90 & 41.34 & 1.57 & 4.19 & 7.73\\
    ROMA & 6.18 & 23.62 & 48.89 & 3.25 & 8.30 & 18.28\\
    \hline
    $\text{Ours}_{\textit{ (ELoFTR)}}$ & 5.93 & 26.72 & \textbf{55.19} & 2.68 & 7.34 & 14.33\\
    $\text{Ours}_{\textit{ (ROMA)}}$ & \textbf{7.78} & \textbf{28.06} & 54.48 & \textbf{4.20} & \textbf{13.08} & \textbf{30.31}\\
    \bottomrule
    \end{tabular}
    }

    \vspace{0.5cm}

    \resizebox{1.0\textwidth}{!}{
    \setlength\tabcolsep{8pt} %
    \begin{tabular}{cccccccccc}
    \toprule
    \multirow{2}{*}{Method}         & \multicolumn{3}{c}{Visible-Thermal Remote Sensing Dataset}             & \multicolumn{3}{c}{Visible-Thermal Aerial View Dataset} & \multicolumn{3}{c}{ Visible-Thermal Ground View Dataset} \\ 
    \cmidrule(lr){2-4}
    \cmidrule(lr){5-7}
    \cmidrule(lr){8-10}
                  & AUC@5pix       & AUC@10pix       & AUC@20pix  & AUC@5$\degree$ & AUC@10$\degree$ & AUC@20$\degree$  & AUC@5$\degree$ & AUC@10$\degree$ & AUC@20$\degree$\\ 
    \midrule
    MCNet & 0.60 & 0.96 & 2.29 & - & - & - & - & - & -\\
    SIFT & 0.07 & 0.08 & 0.09 & 0.00 & 0.00 & 0.04 & 0.00 & 0.00 & 0.00\\
    SRIF & 0.43 & 1.93 & 5.71 & 0.00 & 0.00 & 0.00 & 0.15 & 0.66 & 2.15\\
    SP+LG & 1.55 & 2.83 & 3.84 & 2.93 & 6.65 & 10.80 & 0.82 & 4.01 & 12.17\\
    AspanFormer& 5.00 & 6.78 & 8.74 & 2.38 & 5.70 & 9.21 & 0.95 & 6.73 & 21.29\\
    MatchFormer & 4.69 & 6.56 & 8.35 & 2.05 & 4.66 & 7.82 & 1.12 & 7.42 & 22.41\\
    ELoFTR & 9.33 & 13.21 & 16.31 &  2.13 & 4.93 & 8.67 & 2.34 & 10.86 & 27.36\\ 
    SuperFusion & 0.37 & 0.93 & 1.25 & 0.00 & 0.05 & 0.29 & 0.00 & 0.60 & 3.24 \\
    DKM& 2.61 & 3.81 & 5.36 & 3.37 & 7.69 & 12.28 & 1.79 & 8.94 & 23.28\\
    GIM& 2.33 & 3.35 & 4.02 & 3.46 & 8.43 & 14.47 & 1.22 & 5.93 & 18.23\\
    ROMA& 10.59 & 15.16 & 18.75 & 5.30 & 10.69 & 16.38 & 1.79 & 10.43 & 31.97\\
    \hline
    $\text{Ours}_{\textit{(ELoFTR)}}$ & 9.12 & 21.61 & 36.54 & 4.76 & 9.91 & 15.57 & \textbf{2.38} & 11.43 & 30.93\\
    $\text{Ours}_{\textit{(ROMA)}}$ & \textbf{44.58} & \textbf{57.59} & \textbf{68.26} & \textbf{13.56} & \textbf{25.82} & \textbf{37.23} & 2.32 & \textbf{12.60} & \textbf{36.39}\\
    \bottomrule
    \end{tabular}
    }

    \vspace{0.5cm}

    \resizebox{1.0\textwidth}{!}{
    \setlength\tabcolsep{17pt} %
    \begin{tabular}{cccccccccc}
    \toprule
    \multirow{2}{*}{Method} & \multicolumn{3}{c}{Visible-SAR Dataset} & \multicolumn{3}{c}{Visible-Vectorized Map Dataset} \\ 
    \cmidrule(lr){2-4}
    \cmidrule(lr){5-7}
                  & AUC@5pix       & AUC@10pix       & AUC@20pix  & AUC@5pix       & AUC@10pix       & AUC@20pix\\ 
    \midrule
    MCNet & 0.00 & 0.00 & 0.00 & 0.13 & 0.51 & 3.36\\
    SIFT & 0.00 & 0.00 & 0.00 & 0.00 & 0.00 & 0.00 \\
    SRIF & 0.00 & 0.00 & 0.00 & 0.00 & 0.00 & 0.47\\
    SP+LG& 6.93 & 16.37 & 26.98 & 0.11 & 0.27 & 0.75\\
    AspanFormer & 2.54 & 6.94 & 14.40 & 0.71 & 2.32 & 4.89 \\
    MatchFormer& 5.89 & 15.91 & 29.36 & 5.33 & 17.67 & 32.31\\
    ELoFTR & 3.02 & 9.70 & 21.79 & 2.06 & 10.48 & 23.34\\
    SuperFusion & 0.00 & 0.00 & 0.00 & 4.30 & 25.51 & 56.95\\
    DKM& 7.76 & 16.81 & 27.90 & 0.81 & 2.81 & 3.61\\
    GIM& 10.39 & 23.46 & 38.63 & 0.00 & 0.00 & 0.00\\
    ROMA& 14.96 & 30.29 & 44.54 & 0.00 & 0.00 & 0.00\\
    \hline
    $\text{Ours}_{\textit{ (ELoFTR)}}$ & 15.61 & 37.12 & 61.07 & 10.07 & 38.44 & 60.32 \\
    $\text{Ours}_{\textit{ (ROMA)}}$ & \textbf{29.34} & \textbf{58.05} & \textbf{77.65} & \textbf{21.14} & \textbf{52.31} & \textbf{73.14}\\
    \bottomrule
    \end{tabular}
    }

    \caption{\textbf{Detailed results across multiple datasets using AUC metric.}
    We present statistical results for a full list of methods using the Area Under the Curve (AUC) metric at different thresholds for comparison. 
    The AUC metric provides a more strict assessment than the success rate, as it evaluates overall performance across a broad range of values below a threshold.
    Using AUC metrics, the models trained with our framework still consistently outperform all baselines across all datasets, further demonstrating the effectiveness of our approach. \textbf{Bold} indicates the best performance.
    }
    \label{tab:expauc}
    \end{table*}

\clearpage

\end{document}